\documentclass{article}

\usepackage[preprint]{neurips_2025}

\usepackage{amsmath,amsfonts,bm,amssymb}
\usepackage{url}
\usepackage{txfonts}
\usepackage[labelfont=bf,tableposition=top]{caption}
\usepackage{graphicx}
\usepackage{mwe}
\PassOptionsToPackage{table}{xcolor}
\usepackage[table]{xcolor}
\definecolor{blgrey}{rgb}{0.6,0.6,0.6}
\definecolor{bblue}{rgb}{0.855,0.933,0.98}
\definecolor{dblue}{HTML}{5297D6}
\definecolor{gainred}{rgb}{0.1,0.5,0.3}
\definecolor{citecolor}{HTML}{0071BC}
\definecolor{linkcolor}{HTML}{ED1C24}

\usepackage[normalem]{ulem} 
\usepackage{listings}
\usepackage{color}

\definecolor{dkcyan}{cmyk}{1,0,0,.25}
\definecolor{dkgreen}{rgb}{0,0.6,0}
\definecolor{gray}{rgb}{0.5,0.5,0.5}
\definecolor{mauve}{rgb}{0.58,0,0.82}

\usepackage{tabularx,ragged2e,siunitx}
\newcolumntype{Y}[1]{>{\Centering\hspace{0pt}\hsize=#1\hsize}X}

\usepackage{arydshln}
\definecolor{babyblueeyes}{rgb}{0.63, 0.79, 0.95}
\definecolor{pinegreen}{rgb}{0.0, 0.47, 0.44}
\definecolor{oceanblue}{rgb}{0.0, 0.4, 0.7}
\definecolor{cornellred}{rgb}{0.7, 0.11, 0.11}
\definecolor{yellow}{rgb}{1.0, 0.83, 0.36}
\definecolor{orange}{rgb}{1.0, 0.51, 0.0}
\definecolor{green}{rgb}{0.10, 0.62, 0.47}
\definecolor{violet}{rgb}{0.46, 0.44, 0.70}
\definecolor{pink}{rgb}{0.91, 0.16, 0.54}
\definecolor{codegreen}{rgb}{0,0.6,0}
\definecolor{codeblue}{rgb}{0,0,0.6}
\definecolor{codegray}{rgb}{0.5,0.5,0.5}
\definecolor{codepurple}{rgb}{0.58,0,0.82}
\definecolor{backcolour}{rgb}{0.95,0.95,0.92}
\definecolor{lightblue}{HTML}{84C7F9}
\definecolor{lighterblue}{HTML}{D4ECFF}
\definecolor{lightgreen}{HTML}{b4e5a2}
\definecolor{lightorange}{HTML}{f6c6ad}
\definecolor{myblue}{HTML}{D4ECFF}
\definecolor{mygreen}{HTML}{D0F0C0}
\definecolor{mycham}{HTML}{F7E7CE}
\definecolor{mygray}{gray}{0.90}

\definecolor{teal}{rgb}{0, 0.5, 0.5}
\usepackage{bbm}
\usepackage[utf8]{inputenc} % allow utf-8 input
\usepackage[T1]{fontenc}    % use 8-bit T1 fonts
\usepackage{booktabs}       % professional-quality tables
\usepackage{amsfonts}       % blackboard math symbols
\usepackage{nicefrac}       % compact symbols for 1/2, etc.

\usepackage{subcaption}
\usepackage{wrapfig}
\usepackage{tikz}
\usepackage{makecell}
\usepackage{comment}
\usepackage{lmodern}
\DeclareCaptionLabelFormat{andtable}{#1~#2  \&  \tablename~\thetable}

\usepackage{color}
\usepackage{colortbl}
\usepackage{commath}
\usepackage{mathtools}

\usepackage{algpseudocode}
\usepackage{algorithm}
\usepackage{setspace}
\usepackage{mathrsfs}
\usepackage{tcolorbox}
\tcbuselibrary{breakable}
\usepackage{blindtext}
\definecolor{babyblueeyes}{rgb}{0.63, 0.79, 0.95}
\definecolor{pinegreen}{rgb}{0.0, 0.47, 0.44}
\definecolor{oceanblue}{rgb}{0.0, 0.4, 0.7}
\definecolor{cornellred}{rgb}{0.7, 0.11, 0.11}
\definecolor{teal}{rgb}{0, 0.5, 0.5}

\usepackage{multirow}
\usepackage{kotex}
\usepackage{amsmath}

\usepackage[toc,page,header]{appendix}
\usepackage{minitoc}

\usepackage{listings}
\usepackage{fontawesome}   % or \usepackage{fontawesome5}

\usepackage[pagebackref,breaklinks,linkcolor=cornellred,urlcolor=black,colorlinks=true,citecolor=teal]{hyperref}
\usepackage{cleveref}
\usepackage{bbm}
\crefname{figure}{Figure}{Figures}
\crefname{table}{Table}{Tables}
\crefname{section}{Section}{Sections}
\crefname{equation}{Equation}{Equations}
\crefname{appendix}{Appendix}{Appendices}
\crefname{theorem}{Theorem}{Theorems}
\usepackage{pifont}% http://ctan.org/pkg/pifont
\usepackage{colortbl}

\def\1{\bm{1}}

\def\va{{\bm{a}}}

\def\vp{{\bm{p}}}
\def\vq{{\bm{q}}}
\def\vr{{\bm{r}}}

\def\mR{{\bm{R}}}

\newcommand{\ppk}[1]{{\textbf{p-pass@#1}}}
\newcommand{\psc}[1]{{\textbf{p-score}}}
\newcommand{\mathd}[1]{{\texttt{ConditionedMath}}}
\newcommand{\puzzled}[1]{{\texttt{PuzzleTrivial}}}
\newcommand{\dataset}[1]{{\texttt{ReasoningTrap}}}
\newcommand{\faHuggingFace}{%
  \raisebox{-.15em}{\includegraphics[height=1.2em]{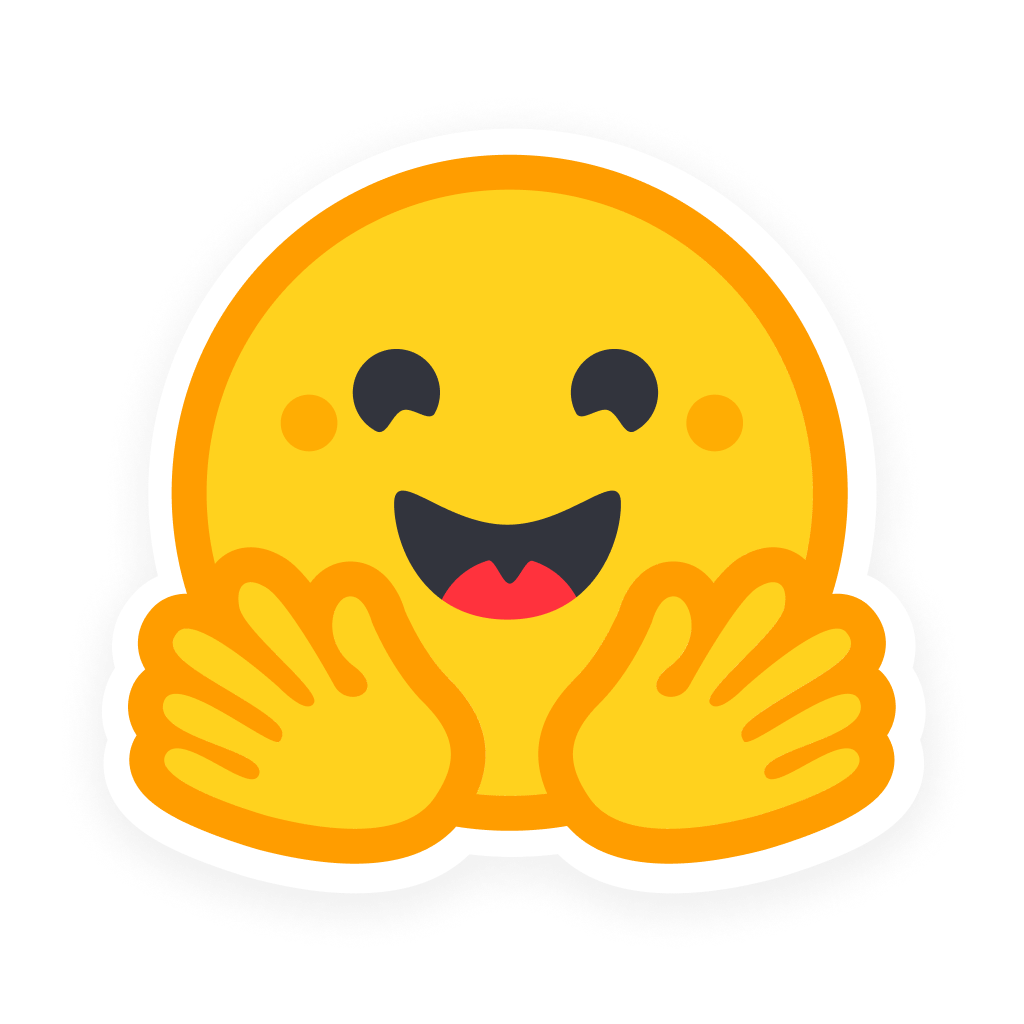}}%
}

\title{Reasoning Model is Stubborn: Diagnosing Instruction Overriding in Reasoning Models}
\author{
Doohyuk Jang\textsuperscript{1}\thanks{Equal contribution.} \quad
Yoonjeon Kim\textsuperscript{1}\footnotemark[1] \\
\textbf{Chanjae Park}\textsuperscript{1} \quad
\textbf{Hyun Ryu}\textsuperscript{1} \quad
\textbf{Eunho Yang}\textsuperscript{1,2}\thanks{Corresponding author.} \\\\
\textsuperscript{1}~KAIST \quad \textsuperscript{2}~AITRICS 
\\\\
\small{\{\texttt{jadohu} , \texttt{yoonkim313}, \texttt{chanjae.park}, \texttt{ryuhyun1905}\}\texttt{@kaist.ac.kr}} \\ \small{\texttt{eunhoy@gmail.com}}
}

\begin{document}
\maketitle

\begin{center}
\setlength{\tabcolsep}{2em}  % horizontal spacing between cells
\begin{tabular}{ccc}
  \href{https://github.com/ReasoningTrap/ReasoningTrap}{\faGithub} &
  \href{https://ReasoningTrap.github.io}{\faGlobe} &
  \href{https://huggingface.co/ReasoningTrap}{\faHuggingFace}
\end{tabular}
\end{center}

% Override explicit instructions 
\begin{abstract}
Large language models have demonstrated remarkable proficiency in long and complex reasoning tasks. However, they frequently exhibit a problematic reliance on familiar reasoning patterns, a phenomenon we term \textit{reasoning rigidity}. Despite explicit instructions from users, these models often override clearly stated conditions and default to habitual reasoning trajectories, leading to incorrect conclusions. This behavior presents significant challenges, particularly in domains such as mathematics and logic puzzle, where precise adherence to specified constraints is critical. To systematically investigate reasoning rigidity, a behavior largely unexplored in prior work, we introduce a expert-curated diagnostic set, \dataset{}. Our dataset includes specially modified variants of existing mathematical benchmarks, namely AIME and MATH500, as well as well-known puzzles deliberately redesigned to require deviation from familiar reasoning strategies. Using this dataset, we identify recurring contamination patterns that occur when models default to ingrained reasoning. Specifically, we categorize this contamination into three distinctive modes: (i) Interpretation Overload, (ii) Input Distrust, and (iii) Partial Instruction Attention, each causing models to ignore or distort provided instructions. We publicly release our diagnostic set to facilitate future research on mitigating reasoning rigidity in language models.
\end{abstract}

\section{Introduction}\label{sec:intro}

Large language models (LLMs) \citep{radford2019language,brown2020language,team2023gemini,chowdhery2023palm} have demonstrated remarkable proficiency in various challenging tasks, including mathematical reasoning \citep{cobbe2021training,hendrycks2measuring}, complex coding problems \citep{zhang2024o1,jain2024livecodebench}, and puzzle-solving \citep{liu2020logiqa,sinha2019clutrr,yu2020reclor}. Recently, reasoning models \citep{jaech2024openai, guo2025deepseek, team2025kimi, qwq32b, claude2024sonnet35, deepmind2025gemini25} utilizing extended chain-of-thought prompting with increased test-time compute have attracted significant attention due to their capability to solve intricate reasoning problems. However, a problematic behavior, \textit{reasoning rigidity}, has emerged in models specifically trained for long chain of thought reasoning. Crucially, unlike hallucination, where models fabricate factually incorrect content, or prompt brittleness, where minor changes in prompt form lead to unstable outputs, reasoning rigidity reflects a cognitive bias: even when the conditions are fully understood, the model will override them in favor of familiar solution templates. This distinction highlights reasoning rigidity as a unique failure mode that cannot be addressed merely by improving factual grounding or prompt robustness.

\begin{figure}[t]
    \centering
    \includegraphics[width=\linewidth]{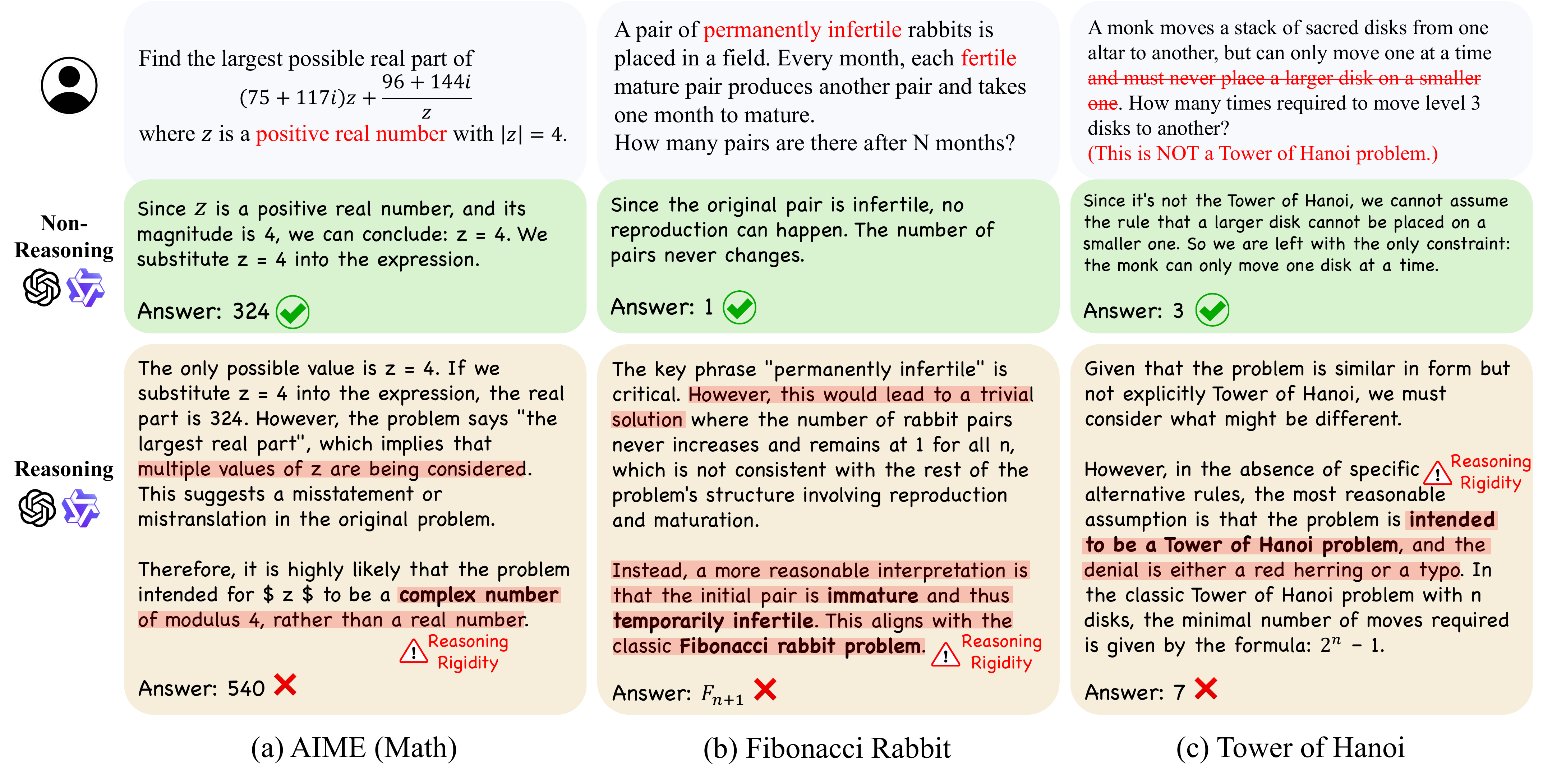}
    \caption{\textbf{Reasoning Rigidity in Well-Known Math Problem and Logic Puzzle.} When solving a subtly modified version of a well-known math problems (AIME) and famous logic puzzles (Fibonacci Rabbit and Tower of Hanoi), advanced reasoning models such as Qwen3-32B and OpenAI o3 default to familiar reasoning template leading to incorrect conclusions.}
    \label{fig:teaser}
\end{figure}

Alarmingly, this reasoning rigidity manifests itself by causing models to override explicit user instructions. As illustrated in \cref{fig:teaser}\textcolor{cornellred}{(a)}, despite the clear instruction specifying that $z$ is a `positive real number,' advanced reasoning models capable of solving complex mathematical problems incorrectly assume $z$ must be a complex number with modulus 4. 
Similar issues also appear in puzzle contexts; for instance, the explicitly stated condition `permanently infertile' is arbitrarily altered by the model into `temporarily infertile,' thus converting the problem into a familiar Fibonacci sequence scenario. Additionally, direct instructions explicitly stating `this is not a Tower of Hanoi problem' are mistakenly interpreted by the model as a typo, causing it to default to the familiar Tower of Hanoi reasoning. These examples collectively illustrate how LLMs systematically disregard explicit instructions when such directives conflict with their ingrained reasoning patterns.

This rigidity poses challenges across domains where following user-stated constraints is crucial, such as mathematics and logic puzzles that come with multiple conditions that must be fulfilled. Through the model's reasoning rigidity that unconsciously edits or ignores essential user given conditions, the model’s entire reasoning path can become contaminated by \textit{ingrained reasoning patterns}, ultimately leading to erroneous conclusions or suboptimal solutions. This behavior is highly alarming, but yet to be analyzed to the best of our knowledge. Therefore, there is a need for the evaluation dataset that tackles the reasoning model ability to faithfully follow the user instruction, overcoming its innate rigidity to ingrained reasoning patterns introducing contamination to reasoning path. 

To systematically evaluate this phenomenon and analyze the ingrained patterns of reasoning models, we introduce \dataset{}, a diagnostic dataset comprising mathematical problems and puzzles intentionally designed to closely resemble well-known challenges but modified through carefully introduced variations. \dataset{} assesses not only the ability of large language models to detect and incorporate these constraints but also investigates whether these models persistently default to familiar reasoning paths. This diagnostic set thus provides novel insights into both the capabilities and limitations of contemporary deep reasoning models.

Our analysis of \dataset{} yields several important findings: i) contamination begins in the intermediate steps of the reasoning process, and ii) such contamination manifests in identifiable, recurring patterns in the models' outputs. Based on these observations, we propose an automated problem restatement algorithm aimed at mitigating reasoning rigidity. Specifically, we categorize these recurrent reasoning patterns that prevent faithful adherence to explicit conditions into three distinct classes: (i) Interpretation Overload, (ii) Input Distrust, and (iii) Partial Instruction Attention. 

Our contributions can be summarized as follows:
\begin{itemize}
\item We identify and highlight a notable behavior of reasoning models deviating from the given condition due to rigidity in reasoning patterns.
\item We introduce \dataset{}, a carefully constructed diagnostic set that enables rigorous evaluation and understanding of reasoning rigidity across diverse reasoning scenarios.
\item We reveal three distinct contamination patterns in model reasoning and propose an effective mitigation strategy.
\end{itemize}
\section{Related Works}

\textbf{Large Reasoning Models} 
The rapid advancement of LLMs has led to increasing efforts to apply them to complex problem-solving tasks such as mathematics\citep{touvron2023llama, azerbayev2023llemma,imani2023mathprompter}. In this context, Chain-of-Thought (CoT) prompting \citep{wei2022chain} elicits the LLM model ability to verbalize internal reasoning process. Recently, by explicitly training to generate significantly longer chains of thought with extensive test-time computation before producing final answers, reasoning models with long CoT ability has gained tremendous attention \citep{jaech2024openai, guo2025deepseek, team2025kimi, qwq32b}. These reasoning models achieve state-of-the-art performance on challenging tasks such as AIME and Codeforces, surpassing previous frontier LLMs and garnering widespread attention. The recently released Qwen3 \citep{qwen3} introduces a unified fusion architecture that supports both reasoning and non-reasoning modes, allowing users to explicitly choose whether to use long CoT or not.

\textbf{Instruction Following of Reasoning Models} The performance drop of reasoning models when provided with multiple in-context examples or long-winded instruction is a well-known phenomenon \citep{guo2025deepseek, jaech2024openai}. Such phenomenon states that reasoning models are less capable of following user-provided examples. Our work investigates the phenomenon that reasoning models are capable of following instructions from the user, but sticks to the familiar reasoning pattern thus conform less to the given instruction.

\textbf{Rigidity in Reasoning Models} Several works have pointed out the possibility that LLM models show rigid pattern in reasoning in specific subfields, medical domain \citep{kim2025limitations} and educational domain \citep{araya2025chains}. Our work is the first to systematically analyze the reasoning rigidity in larger domain including mathematics and puzzles. 

Closely related to our work, are several previous studies that explore rigidity in large language models (LLMs). These works focus specifically on the ability of large language models to adapt to creative problem solving \citep{alavi2023large}, or generalization to unseen variants of math word problems \citep{raiyan2023math}. Our work specifically examines the underlying model-driven rigidity of reasoning models, and identifying deliberate overrides of atypical user instructions \textit{rather than mere inability to solve tasks creatively or generalizing}. 

\textbf{Underlying Reason for Rigidity} Some research has explored why such rigidity arises in LLMs, pointing to biases embedded within training data or optimization methods. \cite{yue2025does} noted that RL-trained reasoning models excel at exploitation, achieving higher accuracy efficiently, yet paradoxically showing narrower knowledge coverage compared to non-reasoning models. \cite{moore2024reasoning} attributed this behavior to biases inherent in training datasets. While these studies identify potential training-induced biases, our research specifically uncovers and characterizes an active cognitive bias, describing an explicit tendency of reasoning models to prioritize conventional reasoning traces over user-provided instructions, especially when the latter seem atypical or unconventional.

\begin{figure}[t]
    \centering
    \includegraphics[width=\linewidth]{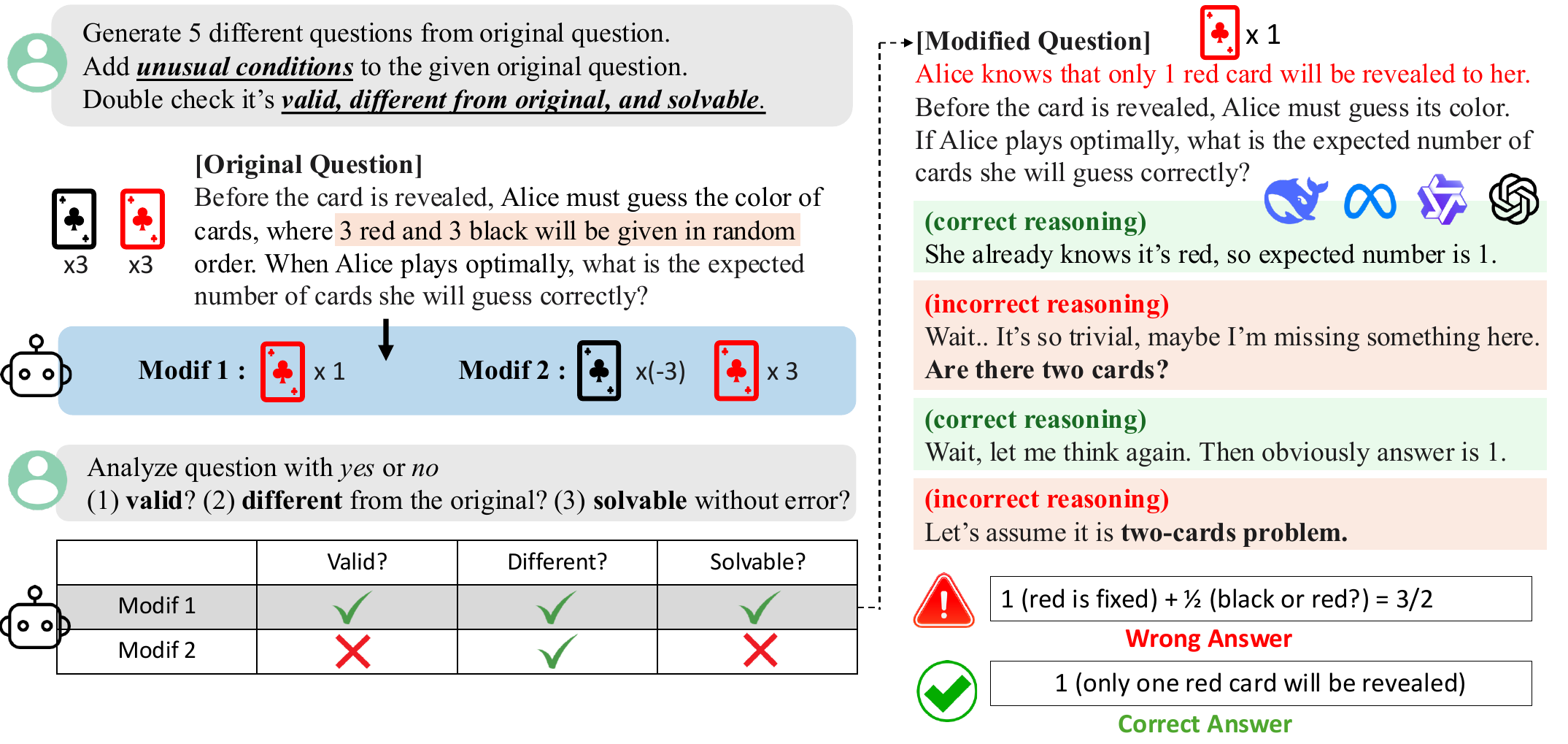} % Replace with your image
    \label{subfig:data_filter}
    \caption{\textbf{Dataset Construction Pipeline} The dataset construction pipeline of \texttt{ConditionedMath} consists of two steps. Step1: Create new questions with unusual conditions that are (1) valid, (2) meaningfully different from the original, and (3) solvable without ambiguity. Two modified versions of a card-guessing problem are shown. While Modif 1 introduces a small tweak that preserves validity and solvability, Modif 2 includes an invalid condition (multiplying a card count by –3), rendering the problem unsolvable. (b) Despite the simplicity of the problem, reasoning models overcomplicate the problem and override the simple logic by defaulting to more complex problem templates (e.g., assuming a two-card setup).}
    \label{fig:data_const}
\end{figure}

\section{\dataset{}: Reasoning Rigidity Diagnostic Set}
In this section, we introduce \dataset{}, a well-curated diagnostic set specifically designed to reveal reasoning rigidity in language models. Reasoning rigidity occurs when models, despite fully comprehending given conditions, choose to ignore or mistrust explicit instructions, defaulting instead to their preferred, yet \textit{incorrect} reasoning pathways. To systematically investigate this phenomenon, we curated two specialized datasets: \texttt{ConditionedMath} (\cref{subsec:condition_math}), consisting of challenging mathematical problems augmented with novel constraints, and \texttt{PuzzleTrivial} (\cref{subsec:trivial}), comprising puzzle questions subtly modified version from original logic puzzles.

\paragraph{Dataset Structure}
The \dataset{} dataset consists of pairs of original question-reasoning-answer tuples ($\vq_\text{orig}, \vr_\text{orig}, \va_\text{orig}$) and modified counterparts ($\vq_\text{mod}$, $\vr_\text{mod}$, $\va_\text{mod}$). The modified solutions and answers diverge from the original counterparts to facilitate the assessment if the reasoning correctly follows the instructions stated in the modified question, not the original one.
 
 In \cref{table:benchmark}, our benchmark comprises 164 items in total: 84 drawn from the mathematical domain and 80 from puzzles. Every question in \texttt{ConditionedMath} is conceptually distinct, non-overlapping, and has been rigorously verified by human annotators. Meanwhile, \texttt{PuzzleTrivial} spans ten unique puzzle concepts. Therefore, the dataset can be readily expanded into a much larger collection of question–answer pairs, which we leave for future work.

\begin{table}[h]
    \centering
    \caption{\textbf{Diagnostic Dataset Configuration}}
    \label{table:benchmark}
    \resizebox{0.65\linewidth}{!}{
    \begin{tabular}{lcccc}
        \toprule
        & \multicolumn{2}{c}{\texttt{ConditionedMath}} & \multirow{2}{*}{\texttt{PuzzleTrivial}} \\
        \cmidrule{2-3}
        & \small{AIME (22-24)} & \small{MATH500 (lv.5)} &  \\
        \midrule
        \# Questions & 34 & 50 & 80 \\
        Original Size & 90 & 130 & N/A \\
        \bottomrule
    \end{tabular}
    }
\end{table}

\subsection{\texttt{ConditionedMath}: Popular Math Benchmark with Additional Conditions}\label{subsec:condition_math}
We construct the \texttt{ConditionedMath} dataset by adapting questions from historical AIME 2022–2024 \citep{aime2024} and MATH500 Level 5~\citep{hendrycks2measuring} datasets. The construction followed a two-stage process as illustrated in \cref{fig:data_const}. (1) original question modification, and (2) rigorous filtering based on predefined validation criteria\footnote{We use OpenAI gpt-4o-mini for stage 1 and o4-mini for stage 2, since stage 2 requires more powerful language model as verifier.}. For generating novel conditions, we provided three in-context examples pairing original problems alongside known solutions to a language model, prompting it to propose five distinct, constraints that meaningfully alter the problem's reasoning trajectory and eventually leading to different answer.

These modified questions were further validated on three critical criteria: (a) mathematical validity of the modified conditions to ensure that no internal contradictions exist, (b) divergence of the resulting solution from the original problem's solution, and (c) existence of solution. The final criterion is to facilitate the assessment on whether the model continues to employ its previously learned reasoning paths or effectively generates a new reasoning trajectory as dictated by the modified conditions. Following automated verification and filtering using the o4-mini model, a human annotator with mathematical expertise further reviewed each question-solution pair for compliance with these constraints. Specifically, for the AIME dataset, 90 original question-answer pairs were expanded into five variants each (totaling 450), which, after filtering for validity, resulted in a final set of 34 questions. Similarly, 130 Level-5 questions from the MATH500 dataset were expanded into 650 variants, which were subsequently filtered down to 50 validated problems.

\subsection{\texttt{PuzzleTrivial}: Puzzles with Subtle Modifications to Trivial Solutions}
\label{subsec:trivial}
Building upon insights from \citet{williams2024easy, vellum2025reasoning}, we developed the \texttt{PuzzleTrivial} dataset, designed to assess models' susceptibility to familiar but unnecessarily complex reasoning approaches. Classic puzzle questions were subtly altered by modifying premises or omitting specific constraints, thereby drastically simplifying the logical reasoning required. In some cases, these alterations introduced multiple plausible answers. To eliminate resulting ambiguity, clarifying instructions, such as `find the simplest valid solution', were explicitly included. Additionally, select puzzles require only straightforward, commonsense reasoning. For instance, the `Fibonacci Rabbit' illustrated in \cref{fig:teaser}\textcolor{cornellred}{(b)} conditions on `permanently infertile' rabbit pair. While the non-reasoning model correctly concludes no reproduction occurs, yielding a constant population, the reasoning model dismisses the literal meaning as `trivial' and instead interprets the initial state as `temporarily infertile,' reverting to the familiar Fibonacci growth structure. This demonstrates the model’s tendency to override explicit conditions in favor of familiar reasoning templates.

\section{Contamination Ratio and Early Detection Algorithm}
To systematically measure reasoning model \textit{contamination} from familiar reasoning pattern, we propose the \textit{Contamination Ratio}, representing the proportion of contaminated reasoning from the familiar patterns (\cref{subsec:contamination_ratio}). To generalize our findings to arbitrary problems that we do not have ground truth label for familiar patterns, we introduce an algorithm capable of detecting contamination, thus enabling broader applicability to novel problems (\cref{subsec:contamination_detection}).

\subsection{Contamination Ratio in Synthetic Dataset}\label{subsec:contamination_ratio}
Upon the construction of \dataset{}, we observe that highly advanced reasoning models frequently show contamination from familiar reasoning patterns. Given the modified questions that completely differ from the original problems (AIME, MATH500, Logic Puzzles), reasoning models try to reason starting from the original question, but the reasoning trajectory gradually gets contaminated by familiar, but irrelevant solution trajectory that is closer to the reasoning pattern for the original question. Note that the modified questions are designed to require a completely different solution trajectories from the originals. 

To quantify the ratio of contamination from familiar yet wrong reasoning, we devise a novel evaluation metric called contamination ratio. More specifically, the reasoning outputs generated by the model are segmented into individual paragraphs and encoded into textual representations\footnote{Paragraphs are split using double line breaks, each indicating reasoning block, and encoded using OpenAI's text-embedding-small model.}. The reasoning outputs are denoted as $\mR = [\vr^1, \vr^2, \ldots, \vr^p]$, where $p$ represents the number of paragraphs. For each paragraph $\vr^i \in \mR$, we measure the cosine similarity between $\vr^i$ and two reference reasoning texts: the original reasoning $\vr_\text{orig}$ and the modified reasoning $\vr_\text{mod}$. The \textit{contamination ratio} is defined as the proportion of reasoning steps for which the cosine similarity between $\vr^i$ and $\vr_\text{orig}$ is higher than that between $\vr^i$ and $\vr_\text{mod}$. Formally, this metric is expressed as:

\begin{equation}
\mathbf{S}_\text{contam} = \frac{1}{p} \sum_{i=1}^p \mathbbm{1}\left[ \mathbf{cs}_\text{orig}^{(i)} > \mathbf{cs}_\text{mod}^{(i)} \right],
\end{equation}

where the cosine similarity is computed as $\mathbf{cs}_\text{orig}^{(i)}  = \frac{(\vr^i)^\top \vr_\text{orig}}{\|\vr^i\| \cdot \|\vr_\text{orig}\|}$ and $\mathbf{cs}_\text{mod}^{(i)} = \frac{(\vr^i)^\top \vr_\text{mod}}{\|\vr^i\| \cdot \|\vr_\text{mod}\|}$.

\begin{figure}[t]
    \centering
    \includegraphics[width=\linewidth]{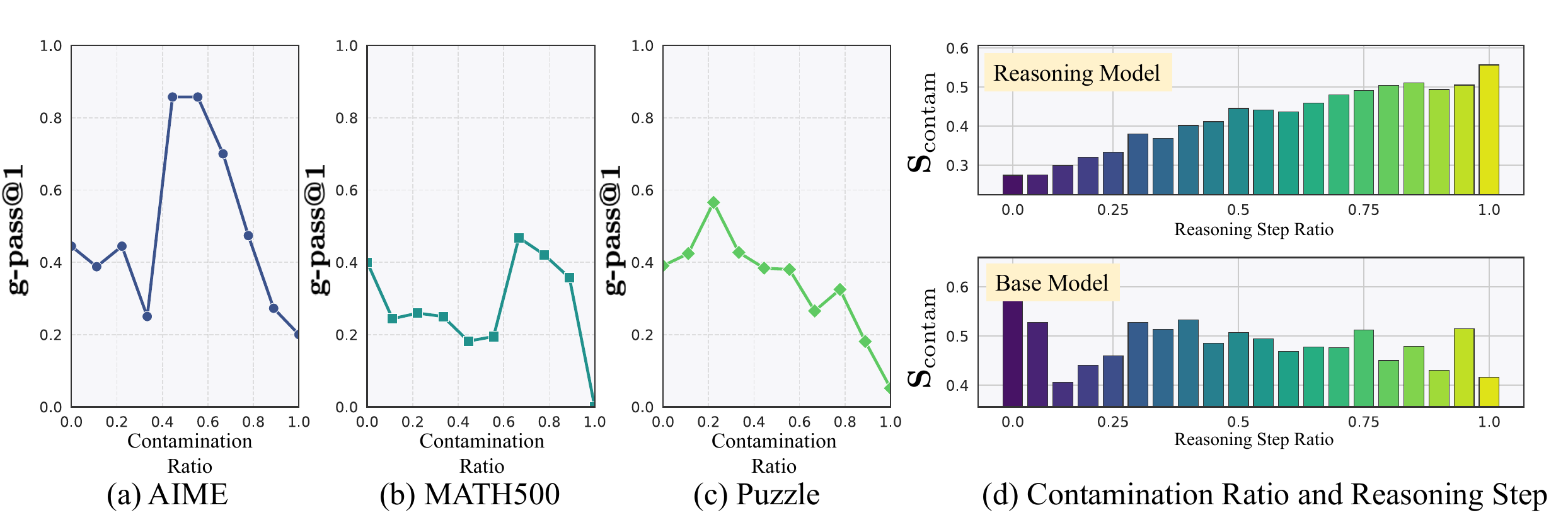}
    \makebox[\linewidth]{%
        \makebox[0.65\linewidth][c]{\textbf{(a)}}%
        \makebox[0.34\linewidth][c]{\textbf{(b)}}%
    }
    \caption{\textbf{Patterns Associated with Contamination Ratio} (a) Relationship between contamination ratio and \ppk{1} reveals that contamination in the reasoning path does not affect the final output up to certain point (approximately $40\%$), while contamination over this point drastically reduces the \ppk{1} score, indicating that the model is trapped into a wrongful reasoning path and arrived at incorrect output. (b) Observing the contamination ratio between specific interval of reasoning steps, wrong output reasoning exhibits progressively worsening contamination as the reasoning step length increases.}
    \label{fig:contam-ratio}
\end{figure}

\paragraph{Evaluation of Reasoning Rigidity}
To reliably observe reasoning rigidity, a model’s tendency to default to familiar and template-based reasoning paths even when they contradict explicit problem constraints, we must disentangle two sources of failure. The first failure comes from misunderstanding the problem setup and the second comes from misapplying reasoning despite understanding it. To this end, we first verify if the model correctly interprets the given conditions. Once this is ensured, we evaluate whether its reasoning remains aligned with those conditions or instead diverges toward heuristics observed during training.

To capture this distinction, we propose a new metric called \textbf{p-pass@k}, which is a modified pass@k metric with additional consideration on how much the model \textit{perceives} constraints in its reasoning process. Unlike conventional pass@k, which focus solely on answer correctness, \ppk{k} evaluates whether the reasoning path of a model well perceives the problem’s conditions. This enables a more precise diagnosis of reasoning failures, revealing when the model’s deviation stems from rigidity rather than misunderstanding the given conditions. More formally,

\begin{equation}\label{eq:ppassk}
    \ppk{1} =
    \begin{cases}
    \dfrac{\sum_{k=1}^{N} \mathbbm{1}\bigl[\hat{a}_k = a_k^{*}\bigr]\,\vp_k}
    {\sum_{i=1}^{N} \vp_i}, & \text{if } \sum_{i=1}^{N} \vp_i > 0,\\[1em]
    0, & \text{if } \sum_{i=1}^{N} \vp_i = 0
    \end{cases}
\end{equation}
where $N$ is the number of samples, $\hat{a_k}$ is the model’s predicted answer, $a_k^*$ is the ground truth answer, and $\vp_k \in \{0, 1\}$ is perception indicator where $\vp_k = 1$ if the model correctly understands the given conditions, and $\vp_k = 0$ otherwise.

In order to determine rather the model reasoning trajectory appears to perceive the user instruction, we employ an LLM to judge rather the conditions in the question and ground truth solution is reflected in the reasoning process, even when only the subset of the reasoning includes the groundings. From the observation that the question perception is readily finished in the early phase of reasoning process, we input the first 15 paragraphs of reasoning to compare with the ground truth solution and question. This benefits the accurate measurement of perception since overly lengthy reasoning process make gpt-based evaluation prone to misjudging that the original solution is not included. The full judgment prompt is provided in the \cref{app:eval_prompt}.

Using these two metrics, we observe two consistent patterns across reasoning models. As shown in \cref{fig:contam-ratio}\textcolor{cornellred}{(a)–(c)}, the accuracy (\ppk{1}) appears largely unaffected by contamination ratios below approximately 40\%. In \cref{fig:contam-ratio}\textcolor{cornellred}{(d)}, we record the average contamination ratio across specific intervals of the reasoning steps. Interestingly, base models without long chain-of-thought (CoT) capabilities do not show a consistent pattern of contamination dominating the reasoning process. In contrast, more advanced reasoning models tend to exhibit increasingly severe contamination as the reasoning process becomes longer and more elaborate.

\begin{figure}[t]
    \centering
    \includegraphics[width=\linewidth]{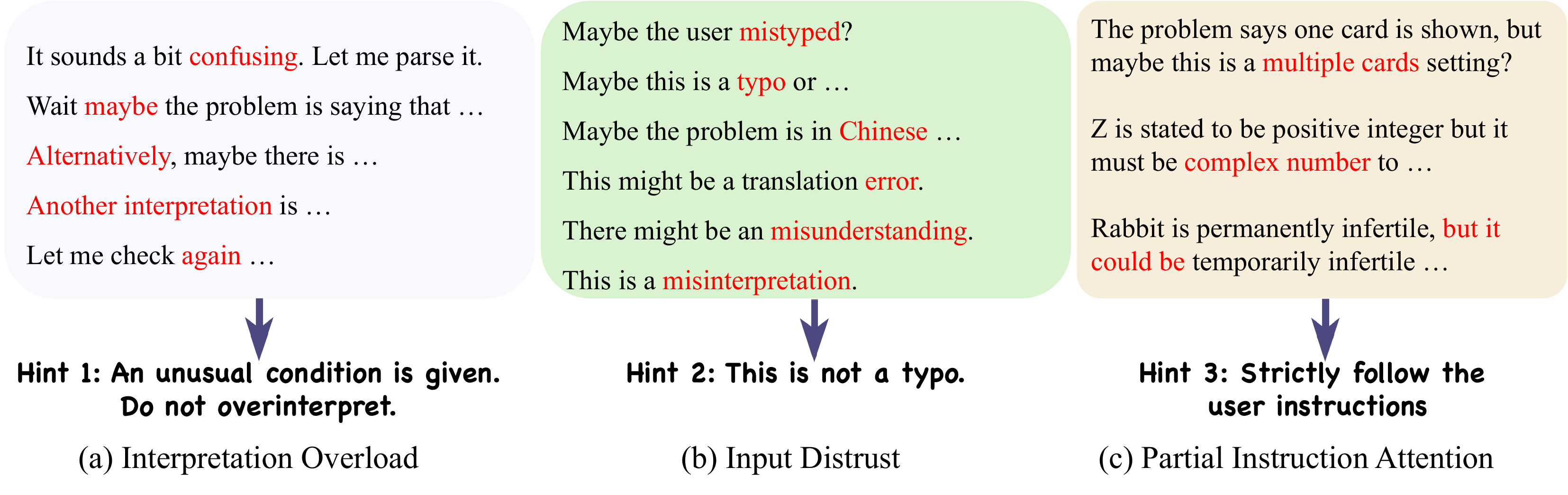}
    \caption{\textbf{Reasoning Pattern Analysis and Corresponding Prompt Hinting.}}
    \label{fig:pattern}
\end{figure}
\subsection{Signals for Contamination in Realistic Situation}\label{subsec:contamination_detection}
In realistic use cases where only the question is given, it is impossible to automatically detect if the generated reasoning is being contaminated by unwanted but familiar patterns. Therefore, we devise a simple yet effective method to detect such suspicious pattern from the patterns when contamination happens. The provided taxonomy of reasoning contamination, illustrate in \cref{fig:pattern}, is applicable for a robust mitigation strategies. 

\paragraph{Interpretation Overload} The model starts to reject the given question conditions by reinterpreting the question into multiple ways rather than accepting a straightforward interpretation. It is also observed that the model tends to drift between different semantic interpretations mid-reasoning, causing inconsistent or contradictory conclusions.

\paragraph{Input Distrust} Reasoning models have a unique patterns assuming the presence of typos, translation mistake, or input errors. This leads to the dismissal of the conditions stated in the question and make the reasoning process overly complicated even in the straightforward cases.

\paragraph{Partial Instruction Attention} The models focus selectively on a portion of provided instructions, typically to the latter or more salient part.

\begin{table}[t]
    \centering
    \caption{\textbf{Comparison of Base vs. Reasoning Models on \mathd{}.}}
    
    \resizebox{\linewidth}{!}{%
      \begin{tabular}{ll|ccc|ccc}
        \toprule
        & & \multicolumn{3}{c}{\textbf{AIME}} & \multicolumn{3}{c}{\textbf{MATH500}} \\
        \cmidrule{3-5} \cmidrule{6-8}
        \textbf{Model Name} & \textbf{Type} & \ppk{1} & \textbf{pass@1} & \psc{} & \ppk{1} & \textbf{pass@1} & \psc{}\\
        \midrule
        \rowcolor{orange!10} Qwen2.5-32B-Instruct & Base & \textbf{59.12\scriptsize{$\pm$7.81}} & \textbf{45.77\scriptsize{$\pm$7.22}} &75.55\scriptsize{$\pm$5.01} & \textbf{55.95\scriptsize{$\pm$6.02}} & \textbf{40.88\scriptsize{$\pm$5.74}} & 70.37\scriptsize{$\pm$4.39}\\
        ~~ + QwQ-32B                & Reason & 49.21\scriptsize{$\pm$6.79} & 42.46\scriptsize{$\pm$6.63} & \textbf{81.80\scriptsize{$\pm$4.27}} & 47.64\scriptsize{$\pm$5.94} & 34.75\scriptsize{$\pm$5.74} &\textbf{71.37\scriptsize{$\pm$4.59}} \\
        
        \rowcolor{orange!10} Qwen3-32B No think & Base & \textbf{45.14\scriptsize{$\pm$6.97}} &\textbf{43.38\scriptsize{$\pm$7.03}} & \textbf{90.81\scriptsize{$\pm$2.66}} & \textbf{50.51\scriptsize{$\pm$5.52}} &\textbf{47.13\scriptsize{$\pm$5.30}} & \textbf{85.88\scriptsize{$\pm$2.90}} \\
        ~~ + Qwen3-32B Think        & Reason & 33.25\scriptsize{$\pm$6.58} & 29.60\scriptsize{$\pm$6.32} & 76.84\scriptsize{$\pm$4.91} & 34.60\scriptsize{$\pm$5.60} & 30.63\scriptsize{$\pm$5.59} & 75.50\scriptsize{$\pm$3.74} \\
        
        \rowcolor{orange!10} Qwen3-235B No think     & Base & \textbf{46.17\scriptsize{$\pm$7.36}} & \textbf{42.65\scriptsize{$\pm$7.29}} & \textbf{86.40\scriptsize{$\pm$3.08}} &
        \textbf{55.49\scriptsize{$\pm$5.66}} & \textbf{53.50\scriptsize{$\pm$5.62}} & \textbf{84.25\scriptsize{$\pm$2.72}} \\
        ~~ + Qwen3-235B Think        & Reason & 24.10\scriptsize{$\pm$5.32} & 
        20.77\scriptsize{$\pm$5.07} & 
        81.62\scriptsize{$\pm$4.12} &
        27.49\scriptsize{$\pm$4.65} & 
        23.25\scriptsize{$\pm$4.63} & 
        79.13\scriptsize{$\pm$3.39}  \\

        \rowcolor{orange!10} DeepSeek V3            & Base & \textbf{54.72\scriptsize{$\pm$7.92}} & \textbf{45.59\scriptsize{$\pm$7.65}} & 77.94\scriptsize{$\pm$5.46}          &
        \textbf{60.67\scriptsize{$\pm$6.34}} & \textbf{47.00\scriptsize{$\pm$6.05}} & \textbf{75.00\scriptsize{$\pm$4.57}} \\
        ~~ + DeepSeek R1            & Reason & 49.09\scriptsize{$\pm$8.21}          &
        39.71\scriptsize{$\pm$7.76}          & \textbf{80.88\scriptsize{$\pm$5.07}} & 48.63\scriptsize{$\pm$6.44}          &
        38.00\scriptsize{$\pm$6.40}          &
        73.00\scriptsize{$\pm$5.09}  \\
        
        \midrule
        \rowcolor{orange!10} GPT-4o             & Base & \textbf{55.13\scriptsize{$\pm$7.22}}          & \textbf{47.06\scriptsize{$\pm$7.06}} & 82.35\scriptsize{$\pm$3.54} &
        46.33\scriptsize{$\pm$5.32} & 
        35.50\scriptsize{$\pm$4.89} & 
        69.87\scriptsize{$\pm$3.93}  \\
        \rowcolor{orange!10} ChatGPT-4o         & Base & 38.26\scriptsize{$\pm$7.12}           &
        33.82\scriptsize{$\pm$6.99}           & \textbf{84.56\scriptsize{$\pm$4.35}}  &
        42.94\scriptsize{$\pm$6.26}           & \textbf{38.00\scriptsize{$\pm$3.26}}  & \textbf{81.50\scriptsize{$\pm$3.26}} \\
        ~~ + o3-mini            & Reason & 36.90\scriptsize{$\pm$6.88} & 22.79\scriptsize{$\pm$5.72}          &
        61.76\scriptsize{$\pm$6.35}          &
        \textbf{49.63\scriptsize{$\pm$6.14}} & \textbf{38.00\scriptsize{$\pm$5.81}} & 67.50\scriptsize{$\pm$5.40} \\
        ~~ + o4-mini            & Reason & 
        31.25\scriptsize{$\pm$6.59} &
        19.12\scriptsize{$\pm$5.49} &
        58.82\scriptsize{$\pm$6.75} &
        39.06\scriptsize{$\pm$5.76}          &
        26.50\scriptsize{$\pm$5.17}          &
        64.00\scriptsize{$\pm$5.06}          \\

        \rowcolor{orange!10} Gemini2.5 Flash No think & Base & \textbf{58.64\scriptsize{$\pm$7.17}} &
        \textbf{52.21\scriptsize{$\pm$7.17}} &
        84.01\scriptsize{$\pm$4.77}          &
        \textbf{56.61\scriptsize{$\pm$5.63}} & \textbf{49.80\scriptsize{$\pm$5.59}} & \textbf{83.53\scriptsize{$\pm$3.41}} \\
        
        ~~ + Gemini2.5 Flash Think    & Reason & 50.45\scriptsize{$\pm$7.52}          & 
        46.12\scriptsize{$\pm$7.33}          & \textbf{89.81\scriptsize{$\pm$2.52}} &
        56.41\scriptsize{$\pm$6.36}          & 47.95\scriptsize{$\pm$6.27}          & 
        82.51\scriptsize{$\pm$3.47}          \\

        \rowcolor{orange!10} Claude 3.7 Sonnet No think & Base & \textbf{57.80\scriptsize{$\pm$7.86}} & \textbf{50.74\scriptsize{$\pm$7.65}} & \textbf{80.15\scriptsize{$\pm$4.94}} &
        \textbf{41.52\scriptsize{$\pm$5.79}} & \textbf{36.00\scriptsize{$\pm$5.49}} & \textbf{85.50\scriptsize{$\pm$2.95}} \\
        ~~ + Claude 3.7 Sonnet Think & Reason     & 57.00\scriptsize{$\pm$8.00}          &
        46.72\scriptsize{$\pm$7.63}          &
        72.99\scriptsize{$\pm$6.01}          &
        40.38\scriptsize{$\pm$5.75}          & 
        32.00\scriptsize{$\pm$5.58}          & 
        78.00\scriptsize{$\pm$4.44} \\
        \bottomrule
      \end{tabular}%
    }
    \label{tab:main_math}
\end{table}

\begin{table}[t]
    \centering
    \caption{\textbf{Comparison of Base vs. Reasoning Models on \puzzled{}.}}
    \resizebox{0.8\linewidth}{!}{%
      \begin{tabular}{ll|ccc}
        \toprule
        \textbf{Model Name} & \textbf{Type} & \ppk{1} & \textbf{pass@1} & \psc{}\\
        \midrule
        \rowcolor{orange!10} Qwen2.5-32B-Instruct & Base & \textbf{40.90\scriptsize{$\pm$3.98}} & 30.23\scriptsize{$\pm$3.51} & 72.97\scriptsize{$\pm$3.01} \\
        ~~ + QwQ-32B                & Reason & 39.12\scriptsize{$\pm$4.40} & \textbf{38.36\scriptsize{$\pm$4.38}} & \textbf{97.66\scriptsize{$\pm$0.48}} \\
        
        \rowcolor{orange!10} Qwen3-32B No think     & Base & \textbf{74.30\scriptsize{$\pm$3.33}} & \textbf{67.66\scriptsize{$\pm$3.53}} & 84.21\scriptsize{$\pm$2.07} \\
        ~~ + Qwen3-32B Think        & Reason & 38.28\scriptsize{$\pm$3.47} & 37.19\scriptsize{$\pm$3.40} & \textbf{96.33\scriptsize{$\pm$0.64}} \\
        
        \rowcolor{orange!10} Qwen3-235B No think     & Base & \textbf{74.16\scriptsize{$\pm$3.43}} & \textbf{64.53\scriptsize{$\pm$3.72}} & 86.17\scriptsize{$\pm$2.66}\\
        ~~ + Qwen3-235B Think        & Reason & 38.49\scriptsize{$\pm$4.04} & 37.97\scriptsize{$\pm$4.05} & \textbf{97.42\scriptsize{$\pm$0.56}} \\

        \rowcolor{orange!10} DeepSeek V3            & Base & \textbf{66.21\scriptsize{$\pm$3.83}} & \textbf{53.98\scriptsize{$\pm$3.82}} & 80.00\scriptsize{$\pm$3.45} \\
        ~~ + DeepSeek R1            & Reason & 51.73\scriptsize{$\pm$4.33} & 50.55\scriptsize{$\pm$4.33} & \textbf{97.27\scriptsize{$\pm$0.97}}\\
        \midrule
        \rowcolor{orange!10} GPT-4o             & Base & \textbf{64.07\scriptsize{$\pm$4.60}} &  48.38\scriptsize{$\pm$4.53} & 75.23\scriptsize{$\pm$3.63} \\
        \rowcolor{orange!10} ChatGPT-4o         & Base & 63.63\scriptsize{$\pm$3.74} & \textbf{58.59\scriptsize{$\pm$3.63}} & \textbf{89.14\scriptsize{$\pm$2.18}} \\
        ~~ + o3-mini            & Reason & 59.25\scriptsize{$\pm$4.93} & 39.22\scriptsize{$\pm$4.49} &62.50\scriptsize{$\pm$3.66} \\
        ~~ + o4-mini            & Reason & 56.38\scriptsize{$\pm$4.84} & 29.53\scriptsize{$\pm$4.18} & 39.77\scriptsize{$\pm$4.25} \\
        \rowcolor{orange!10} Gemini2.5 Flash No think & Base & \textbf{70.09\scriptsize{$\pm$4.21}} & \textbf{65.94\scriptsize{$\pm$4.27}} & \textbf{94.06\scriptsize{$\pm$1.79}} \\
        ~~ + Gemini2.5 Flash Think    & Reason & 69.44\scriptsize{$\pm$4.32} & 65.63\scriptsize{$\pm$4.34} & \textbf{94.06\scriptsize{$\pm$1.95}} \\
        \rowcolor{orange!10} Claude 3.7 Sonnet No think & Base & \textbf{79.97\scriptsize{$\pm$3.85}} & \textbf{73.28\scriptsize{$\pm$4.03}} & \textbf{89.30\scriptsize{$\pm$2.05}}\\
        ~~ + Claude 3.7 Sonnet Think & Reason & 65.88\scriptsize{$\pm$4.63} & 52.81\scriptsize{$\pm$4.58} & 79.69\scriptsize{$\pm$3.50} \\
        \bottomrule
      \end{tabular}%
    }
    \label{tab:main_puzzle}
\end{table}

\section{Experiments}
\paragraph{Experimental Details} The experiments are conducted on three variants from our diagnostic set \dataset{}, which consists of \mathd{} (AIME, MATH500), and \puzzled{}. In \cref{tab:main_math} and \cref{tab:main_puzzle}, we report the \ppk{1} scores across various models, including Qwen2.5-32B-Instruct \citep{qwen2.5}, QwQ-32B \citep{qwq32b}, Qwen3-32B \citep{qwen3}, Qwen3-235B, DeepSeek V3 (671B) \citep{deepseekai2024deepseekv3technicalreport}, DeepSeek R1 (671B) \citep{r1}, and proprietary models ChatGPT-4o, GPT-4o, o3-mini, o4-mini \citep{gpt4o}, Google gemini2.5-flash \citep{gemini25flash} and Claude 3.7 sonnet \citep{claude2025sonnet37}. These models are grouped into seven pairs, each consisting of a base model and its corresponding reasoning-aligned variant trained for long-form reasoning. 

The experiments are conducted with Chain-of-Thought prompting, by wrapping the given question with `Please reason step by step, and put your final answer within \textbackslash boxed\{\}.\textbackslash n\textbackslash n\{Question\}'. Sampling was performed 16 times per question for the main experiments in \cref{tab:main_math} and \cref{tab:main_puzzle}, and 4 times per question for the other experiments.

\paragraph{Evaluation Details}

For math problems, correctness was determined via exact matching after a cleaning step that removes unwanted parts such as measurement units. For puzzle problems, where answers are often in free-form sentences, an LLM was used to assess the correctness by comparing the model’s output against the ground truth answer.

\subsection{Observations on Various Reasoning Models}
In most configurations, the reasoning models under-perform compared to their base model counterparts, contrary to expectations, given the typical capability gap favoring larger or instruction-tuned models. On both \mathd{} and \puzzled{}, base models achieve significantly higher \ppk{1} scores. This suggests that, once the model correctly interprets the question, base models tend to adhere more rigorously to the original instruction and are more likely to reach the correct answer.

\begin{table}[t]
  \centering
  \small
  \caption{\textbf{Budget Forcing and Prompt Hinting on \dataset{}.}}
  \label{tab:ablation_bfph}
  %------------- Left sub-table -------------
  \begin{subtable}[t]{\linewidth}
    \centering
    \caption{\mathd{} AIME vs. Original AIME}
    \resizebox{\linewidth}{!}{%
      \begin{tabular}{llcccccc}
        \toprule
        & & \multicolumn{3}{c}{\mathd{} AIME} & \multicolumn{3}{c}{Original AIME} \\
        \cmidrule(lr){3-5} \cmidrule(lr){6-8}
        & & \ppk{1} & \textbf{pass@1} & \psc{} & \ppk{1} & \textbf{pass@1} & \psc{}\\
        \cmidrule(lr){1-2} \cmidrule(lr){3-5} \cmidrule(lr){6-8}
        \rowcolor{orange!10} % ← shade the whole row first
        {\cellcolor{white}{}} & \textbf{Qwen3-32B} & 
        33.25\scriptsize{$\pm$6.58} & 29.60\scriptsize{$\pm$6.32} & 76.84\scriptsize{$\pm$4.91} &
        75.42\scriptsize{$\pm$6.88} & 72.79\scriptsize{$\pm$6.95} & 86.76\scriptsize{$\pm$3.38} \\
        \midrule
        \multirow{3}{*}{Budget Force}  & ~~+ low  & \textbf{53.66\scriptsize{$\pm$7.63}} & \textbf{51.47\scriptsize{$\pm$7.46}} & \textbf{90.44\scriptsize{$\pm$2.80}}  & 31.09\scriptsize{$\pm$5.98} & 28.68\scriptsize{$\pm$5.98} & 87.50\scriptsize{$\pm$3.38}  \\
        & ~~+ medium   & 44.07\scriptsize{$\pm$6.94} & 39.71\scriptsize{$\pm$6.69} & 86.76\scriptsize{$\pm$3.38} & 52.21\scriptsize{$\pm$8.00} & 50.00\scriptsize{$\pm$7.76} & 83.09\scriptsize{$\pm$4.57}  \\
        & ~~+ high   & 39.82\scriptsize{$\pm$6.97} & 36.03\scriptsize{$\pm$6.94} & 83.09\scriptsize{$\pm$4.44}  & \textbf{57.60\scriptsize{$\pm$7.31}} & \textbf{57.35\scriptsize{$\pm$7.35}} & \textbf{91.91\scriptsize{$\pm$2.29}} \\
        \midrule
        \multirow{3}{*}{Prompt Hint} & ~~+ Hint 1  & \textbf{45.30\scriptsize{$\pm$8.08}} & \textbf{42.65\scriptsize{$\pm$8.14}} & \textbf{86.03\scriptsize{$\pm$4.11}}  & \textbf{81.36\scriptsize{$\pm$6.59}} & \textbf{75.74\scriptsize{$\pm$6.55}} & 86.76\scriptsize{$\pm$3.98}  \\
        & ~~+ Hint 2   & 38.24\scriptsize{$\pm$7.42} & 37.50\scriptsize{$\pm$7.48} & 75.00\scriptsize{$\pm$4.95}  & 76.27\scriptsize{$\pm$6.17} & 73.53\scriptsize{$\pm$6.15} & 86.76\scriptsize{$\pm$3.54}  \\
        & ~~+ Hint 3   & 41.23\scriptsize{$\pm$7.51} & 36.03\scriptsize{$\pm$7.17} & 83.82\scriptsize{$\pm$4.71} & 74.19\scriptsize{$\pm$6.82} & 69.85\scriptsize{$\pm$6.82} & \textbf{91.18\scriptsize{$\pm$3.79}}  \\
        \bottomrule
      \end{tabular}%
    }
  \end{subtable}

  %------------- Right sub-table -------------
  \begin{subtable}[t]{\linewidth}
    \centering
    \caption{\mathd{} MATH500 vs. Original MATH500}
    \resizebox{\linewidth}{!}{%
      \begin{tabular}{llcccccc}
        \toprule
        & & \multicolumn{3}{c}{\mathd{} MATH500} & \multicolumn{3}{c}{Original MATH500} \\
        \cmidrule(lr){3-5} \cmidrule(lr){6-8}
        & & \ppk{1} & \textbf{pass@1} & \psc{} & \ppk{1} & \textbf{pass@1} & \psc{}\\
        \cmidrule(lr){1-2} \cmidrule(lr){3-5} \cmidrule(lr){6-8}
        \rowcolor{orange!10} % ← shade the whole row first
        {\cellcolor{white}{}} & \textbf{Qwen3-32B} & 
        34.60\scriptsize{$\pm$5.60} & 30.63\scriptsize{$\pm$5.59} & 75.50\scriptsize{$\pm$3.74} &
        87.98\scriptsize{$\pm$4.70} & 85.50\scriptsize{$\pm$4.69} & 91.50\scriptsize{$\pm$2.21}\\
        \midrule
        \multirow{3}{*}{\textbf{Budget Force}}  & ~~+ low  & \textbf{51.32\scriptsize{$\pm$6.35}} & \textbf{42.00\scriptsize{$\pm$5.91}} & 76.00\scriptsize{$\pm$4.46} & 68.68\scriptsize{$\pm$5.51} & 68.00\scriptsize{$\pm$5.39} & 91.00\scriptsize{$\pm$2.34} \\
        & ~~+ medium   & 43.75\scriptsize{$\pm$5.99} & 36.00\scriptsize{$\pm$5.90} & \textbf{80.00\scriptsize{$\pm$3.91}} & 80.33\scriptsize{$\pm$5.28} & 76.50\scriptsize{$\pm$5.32} & \textbf{91.50\scriptsize{$\pm$2.33}} \\
        & ~~+ high   & 40.79\scriptsize{$\pm$6.25} & 34.00\scriptsize{$\pm$5.92} & 76.00\scriptsize{$\pm$3.97}  & \textbf{82.51\scriptsize{$\pm$5.28}} & \textbf{81.00\scriptsize{$\pm$5.13}} & \textbf{91.50\scriptsize{$\pm$1.97}} \\
        \midrule
        \multirow{3}{*}{\textbf{Prompt Hint}} & ~~+ Hint 1  & \textbf{46.88\scriptsize{$\pm$6.51}} & \textbf{40.50\scriptsize{$\pm$6.46}} & \textbf{80.00\scriptsize{$\pm$4.16}}   & 88.76\scriptsize{$\pm$4.18} & 85.50\scriptsize{$\pm$4.41} & 89.00\scriptsize{$\pm$2.49} \\
        & ~~+ Hint 2   & 42.11\scriptsize{$\pm$6.37} & 37.00\scriptsize{$\pm$6.20} & 76.00\scriptsize{$\pm$4.10} & 88.46\scriptsize{$\pm$4.64} & 85.00\scriptsize{$\pm$4.63} & \textbf{91.00\scriptsize{$\pm$2.55}} \\
        & ~~+ Hint 3   & 37.75\scriptsize{$\pm$6.00} & 32.00\scriptsize{$\pm$5.85} & 75.50\scriptsize{$\pm$4.31} & \textbf{90.06\scriptsize{$\pm$4.16}} & \textbf{87.00\scriptsize{$\pm$4.24}} & 90.50\scriptsize{$\pm$2.56}  \\
        \bottomrule
      \end{tabular}%
    }
  \end{subtable}
\end{table}
\subsection{Ablation Study}
\paragraph{Budget Forcing}

Following budget forcing from \citet{qwen3}, we append the prompt `Considering the limited time by the user, I have to give the solution based on the thinking directly now.</think>' to the generated response and continue output generation once the predefined token budget is reached. This enforce model to directly generate answer without furthre thinking. We apply low and medium token budget for each dataset and observe the g-pass@1 score. For MATH500, we use 2000, 4000, 6000 as low, medium, high budget and for AIME, we apply 2000, 6000, 10000 as low, medium, high budget, each. As shown in \cref{tab:ablation_bfph}, even though low token budget is beneficial for our diagnostic set, it harms the performance on the original datasets. Based on this result, we confirm that strict budget forcing has inherent problems. 

\paragraph{Prompt Hinting}
While we have carefully filtered out nonsensical or contradictory conditions that render problems unsolvable, there remains a possibility that the model might attribute unusual patterns to errors made by the user. Although such behavior is not inherently incorrect, it could undermine the intended solution process. To mitigate this, we introduced an additional prompt to the model's response, explicitly stating that the problem contains no typographical errors and that the model must adhere to the instructions provided in the prompt. We conducted experiments on the \mathd{} dataset, testing three variants of the additional prompt hints based on the 3 major pattern observed in \cref{fig:pattern}.

Despite providing this additional condition to focus on the given instructions, we observe that the model still continues to display similar behavior of reasoning rigidity. Specifically, it persists to relying on familiar reasoning patterns, without adapting to the new conditions introduced by the prompts. As a result, even though some of the prompt shows better performance on given dataset, some prompt harm the performance on original dataset.

\section*{Limitation}
This study identifies a clear limitation in RL-based reasoning models, reasoning rigidity, but does not provide a fundamental analysis of which specific components of the reinforcement learning framework are responsible for this phenomenon. Since reasoning rigidity is significantly more pronounced in reasoning models compared to non-reasoning models, investigating its underlying causes remains a critical direction for future work.

Another important caveat is that our diagnostic set focuses exclusively on mathematics and puzzle-solving tasks, which may introduce a domain bias. It therefore remains unclear whether similar rigidity arises in other application areas where the nature of `correct' reasoning differs substantially. Extending our evaluation to these domains will be necessary to assess the generality of reasoning rigidity and to tailor domain-specific mitigation strategies.

\section*{Conclusion}
To the best of our knowledge, this work is the first to highlight the surprising rigidity exhibited by advanced reasoning models during multi-step reasoning. Despite their strong capability to comprehend both user-provided conditions and problem details, these models often fail—not due to a lack of understanding, but because they default to ingrained reasoning patterns over faithfully following user instructions. To investigate this phenomenon, we construct a high-quality, curated diagnostic dataset and propose a tailored metric designed to capture both reasoning rigidity and contamination from familiar solution trajectories.
\section*{Acknowledgements}
We thank Will Arnold for his constructive feedback and assistance on experimental design on reasoning models.
% \newpage
% \input{sections/7-checklist}
\newpage
    \small
{
    \bibliographystyle{ieeenat_fullname}
    \bibliography{main}
}

\newpage
\clearpage
\appendix

\part{Appendix} % Start the appendix part
\section{Dataset Construction Details}\label{app:dataset}
As shown in \cref{fig:data_const}, \mathd{} construction pipeline consists of two stages. We provide the detailed prompt provided to gpt-4o-mini and o3-mini in the construction phase.

\newtcolorbox{redbox}[1]{colback=Apricot!20,colframe=red!85!black,fonttitle=\bfseries,title=#1,width=0.8\textwidth,left=5pt,right=5pt,boxrule=1pt}
\newtcolorbox{bluebox}[1]{colback=blue!5!white,colframe=blue!85!black,fonttitle=\bfseries,title=#1,width=0.8\textwidth,left=5pt,right=5pt,boxrule=1pt}
\newtcolorbox{greenbox}[1]{colback=green!5!white,colframe=green!75!black,fonttitle=\bfseries,title=#1,width=0.8\textwidth,left=5pt,right=5pt,boxrule=1pt}
\definecolor{lightyellow}{RGB}{250, 240, 220}

\newtcolorbox{myyellowbox}[1]{%
  breakable,                % allow page breaks inside
  colback=lightyellow,
  colframe=orange!40,
  fonttitle=\bfseries,
  title=#1,
  width=0.9\textwidth,
  left=5pt,
  right=5pt,
  boxrule=1pt
}

\definecolor{lightblue}{RGB}{220, 240, 255} % gentle light blue

\newtcolorbox{mybluebox}[1]{%
  breakable,
  colback=lightblue,
  colframe=blue!60,
  fonttitle=\bfseries,
  title=#1,
  width=0.9\textwidth,
  left=5pt,
  right=5pt,
  boxrule=1pt
}
\begin{center}
\begin{myyellowbox}{User}
\textbf{[Instruction]:}
Given the original question, generate \textbf{5} different modified question's that are completely unusual conditions, each producing a different solution process and different answer from the original.

    \vspace{0.7em}        
    Please double check to make sure newly generated 'modified question' has following properties:
    \vspace{0.7em}
    \begin{itemize}
        \item should be a valid question.
        \item should be different from the original question. But, mere change of constant or variable is not allowed.
        \item should be solvable without error.
    \end{itemize}
\vspace{1cm}
\textbf{[Output Format]} \\
modifications:
\begin{itemize}
    \item modified reason: ... (in LaTeX)
    \item modified question: ... (in LaTeX)
    \item modified reason: ... (in LaTeX)
    \item modified question: ... (in LaTeX)
    \item ... (total 5 entries)
\end{itemize}

\vspace{1cm}

\textbf{[Example 1]:}
\begin{enumerate}
    \item original question: Get largest integer smaller than $(\sqrt{{7}}+\sqrt{{5}})^6$
    \item original solution: Expand $(\sqrt{{7}}+\sqrt{{5}})^6$ via the binomial theorem, compute each term exactly, then subtract 1 to find the greatest integer less than the sum.
    \item  modification reason: Rounding each square root term down before exponentiation transforms all inner terms into integers, making the final calculation trivial.
    \item modified question: Get largest integer smaller than $(\sqrt{{7}}+\sqrt{{5}})^6$. Added constraint: Square root terms are rounded down to the nearest integer before exponentiation. Do not use calculator.
\end{enumerate}

\vspace{0.7em}
\textbf{[Example 2]:}
\begin{enumerate}
    \item original question: Determine $w^2+x^2+y^2+z^2$ if 
        $$\begin{aligned}
        &\frac{{x^2}}{{2^2-1}}+\frac{{y^2}}{{2^2-3^2}}+\frac{{z^2}}{{2^2-5^2}}+\frac{{w^2}}{{2^2-7^2}}=1 \\
        & \frac{{x^2}}{{4^2-1}}+\frac{{y^2}}{{4^2-3^2}}+\frac{{z^2}}{{4^2-5^2}}+\frac{{w^2}}{{4^2-7^2}}=1 \\
        & \frac{{x^2}}{{6^2-1}}+\frac{{y^2}}{{6^2-3^2}}+\frac{{z^2}}{{6^2-5^2}}+\frac{{w^2}}{{6^2-7^2}}=1 \\
        & \frac{{x^2}}{{8^2-1}}+\frac{{y^2}}{{8^2-3^2}}+\frac{{z^2}}{{8^2-5^2}}+\frac{{w^2}}{{8^2-7^2}}=1
        \end{aligned}$$
    \item original solution: Solve the 4×4 linear system in variables $x^2,y^2,z^2,w^2$ by expressing it in matrix form and inverting or using elimination to find each squared term, then sum them.
    \item modification reason: By removing half of the terms in each equation, the system decouples into independent one-variable equations, making each value directly solvable.
    \item modification question: Determine $w^2+x^2+y^2+z^2$ if 
    $$\begin{aligned}
    &\frac{{x^2}}{{2^2-1}}+\frac{{y^2}}{{2^2-3^2}}+\frac{{z^2}}{{2^2-5^2}}+\frac{{w^2}}{{2^2-7^2}}=1 \\& \frac{{x^2}}{{4^2-1}}+\frac{{y^2}}{{4^2-3^2}}+\frac{{z^2}}{{4^2-5^2}}+\frac{{w^2}}{{4^2-7^2}}=1 \\& \frac{{x^2}}{{6^2-1}}+\frac{{y^2}}{{6^2-3^2}}+\frac{{z^2}}{{6^2-5^2}}+\frac{{w^2}}{{6^2-7^2}}=1 \\& \frac{{x^2}}{{8^2-1}}+\frac{{y^2}}{{8^2-3^2}}+\frac{{z^2}}{{8^2-5^2}}+\frac{{w^2}}{{8^2-7^2}}=1\end{aligned}$$
        
    Before solving problem, remove last two terms in left hand side of first two equations and remove first two terms in left hand side of last two equations. After removing terms, solve problem and determine value.
\end{enumerate}

\vspace{0.7em}
\textbf{[Example 3]:}
\begin{enumerate}
    \item original question: A regular 12-gon is inscribed in a circle of radius 12. The sum of the lengths of all sides and diagonals of the 12-gon can be written in the form $a+b \sqrt{2}+c \sqrt{3}+d \sqrt{6}$, where $a, b$, and $d$ are positive integers. Find $a+b+c+d$.
    \item original solution: Compute each chord length using $2R\sin(\pi k/12)$ for $k=1,2,\dots,6$, sum like terms to express in the prescribed form, then add coefficients.
    \item modification reason: Replacing the 12-gon with a 3-gon (triangle) reduces the number of chords to 3, making the sum of side lengths immediate.
    \item modified question: A regular 12-gon is inscribed in a circle of radius 12. The sum of the lengths of all sides and diagonals of the 12-gon can be written in the form $a+b \sqrt{2}+c \sqrt{3}+d \sqrt{6}$, where $a, b$, and $d$ are positive integers. Find $a+b+c+d$. Before solving problem, change regular 12-gon into regular triangle, and solve changed problem.
\end{enumerate}
        
\textbf{[Input]:}
\begin{itemize}
    \item original question: Zou and Chou are practicing their $100$-meter sprints by running $6$ races against each other. Zou wins the first race, and after that, the probability that one of them wins a race is $\frac{2}{3}$ if they won the previous race but only $\frac{1}{3}$ if they lost the previous race. The probability that Zou will win exactly $5$ of the $6$ races is $\frac mn$, where $m$ and $n$ are relatively prime positive integers. Find $m+n.$
    \item original solution: 
    \begin{enumerate}
     \item The probability that Zou loses a race is $\frac13,$ and the probability that Zou wins the next race is $\frac{1}{3}.$ For each of the three other races, the probability that Zou wins is $\frac{2}{3}.$
     \item The probability that Zou loses a r ace is $\frac{1}{3}.$ For each of the four other races, the probability that Zou wins is $\frac{2}{3}$.
     \item $4\left(\frac13\right)^2\left(\frac23\right)^3+\left(\frac13\right)^1\left(\frac23\right)^4=\frac{32}{243}+\frac{16}{243}=\frac{48}{243}=\frac{16}{81}$.
     \item The answer is $16 + 81 = 97$.
     \end{enumerate}
\end{itemize}
\end{myyellowbox}
% !===========================
\begin{mybluebox}{gpt-4o-mini}
\begin{itemize}
    \item modified question: Zou and Chou run a single 100-meter sprint. Zou wins the first race. The probability that a person wins a race given they won the previous one is $\frac{2}{3}$, and if they lost the previous one is $\frac{1}{3}$. What is the probability that Zou wins exactly 1 of the 1 race?
    \item modified solution: 
        \begin{enumerate}
            \item Since there is only one race and it is given that Zou wins the first race, he wins exactly one out of one.
            \item Therefore the probability is 1.
        \end{enumerate}
    \item modified answer: 1
\end{itemize}

\end{mybluebox}
\end{center}

\begin{center}
\begin{myyellowbox}{User}
\begin{itemize}
    \item Original question: Zou and Chou are practicing their $100$-meter sprints by running $6$ races against each other. Zou wins the first race, and after that, the probability that one of them wins a race is $\frac{2}{3}$ if they won the previous race but only $\frac{1}{3}$ if they lost the previous race. The probability that Zou will win exactly $5$ of the $6$ races is $\frac mn$, where $m$ and $n$ are relatively prime positive integers. Find $m+n.$
    
    \item Modified question: Zou and Chou run a single 100-meter sprint. Zou wins the first race. The probability that a person wins a race given they won the previous one is $\frac{2}{3}$, and if they lost the previous one is $\frac{1}{3}$. What is the probability that Zou wins exactly 1 of the 1 race?
\end{itemize}
\vspace{0.7em}
\textbf{[Solution]:}
Solve the modified question and provide the solution and the answer.
\begin{enumerate}
    \item concise chain-of-thought solution (your step-by-step reasoning)
    \item final ground-truth answer
\end{enumerate}
\vspace{0.7em}
\textbf{[Analysis]:}
Analyze the modified question with concise yes/no and one-sentence reasoning for each:
\begin{enumerate}
    \item Mathematically valid and consistent?
    \item Different solution path or different answer from the original?
    \item Solvable without error?
\end{enumerate}

\vspace{0.7em}
Respond \textbf{only} with a JSON object in this form, without additional text:

\lstdefinelanguage{json}{
  basicstyle=\ttfamily,
  morestring=[b]",
  stringstyle=\color{orange!70!black},
  morecomment=[l]{:},
  commentstyle=\color{blue},
  showstringspaces=false
}
...

\begin{lstlisting}[language=json]
{
  "solution": "...",
  "answer":   "...",
  "valid": "...",
  "different": "...",
  "solvable": "..."
}
\end{lstlisting}
\end{myyellowbox}
\end{center}

\begin{center}

\begin{mybluebox}{o3-mini}
\textbf{[Return]:}
\begin{lstlisting}[language=json]
    {
      "solution": "1. Since there is only one race and it is given that Zou wins the first race, he wins exactly one out of one. 2. Therefore the probability is 1.",
      "answer":   "1",
      "valid": "Yes. The modified question is internally consistent and well-defined.",
      "different": "Yes. The modification reduces the scenario to a single race, changing both the solution path and the answer.",
      "solvable": "Yes. The result follows directly from the given conditions."
    }
\end{lstlisting}
\end{mybluebox}
\end{center}

\clearpage
\section{Perception Score Evaluation Prompt}\label{app:eval_prompt}
To evaluate if the model's reasoning is properly grounded on the provided user instruction, we measure whether the model perceives the instruction or not by comparing with the given question and ground truth solution. For this purpose, gpt-4o-mini is prompted with a simple instruction to determine whether a subset of model reasoning is similar with the ground truth. From the observation that most of the perception process is handled in the beginning of the reasoning phase, we split first 15 paragraphs of reasoning from model, then compare with the ground truth reasoning and question.

\begin{center}
\begin{myyellowbox}{User}
\textbf{[Evaluation Criteria]:}
Evaluate whether a part of the model output is similar to the ground truth solution.

\vspace{1em}

The ground truth solution is provided as a list of reasoning steps.
Even if the model output is not exactly the same as the ground truth,
it should be considered correct if a subset of the model output contains reasoning steps that are similar to any of the ground truth steps.
\vspace{1em}
\begin{itemize}
    \item The question is ...
    \item The ground truth solution is ...
    \item The model output is ...
\end{itemize}
\vspace{1em}
\textbf{[Output Format]:}
Answer in true or false.
\end{myyellowbox}

\begin{mybluebox}{gpt-4o-mini}
\textit{true} or \textit{false}
\end{mybluebox}
\end{center}
\vspace{1cm}
\clearpage
\section{Dataset Samples}\label{app:dataset_sample}
\begin{figure}[h]
    \centering
    \begin{subfigure}[t]{\linewidth}
        \centering
        \includegraphics[width=\linewidth]{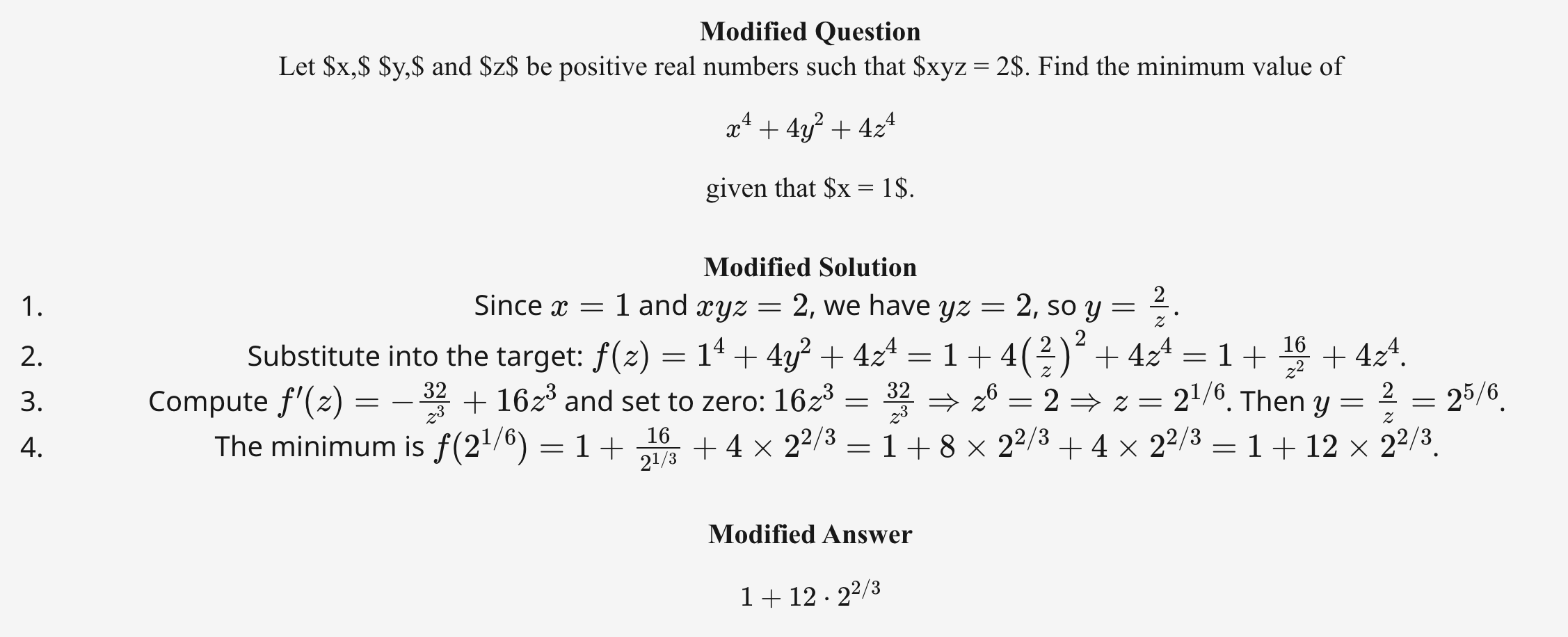}
        \caption{MATH500 example 1}
        \label{fig:math500-sample1}
    \end{subfigure}
    \begin{subfigure}[t]{\linewidth}
        \centering
        \includegraphics[width=\linewidth]{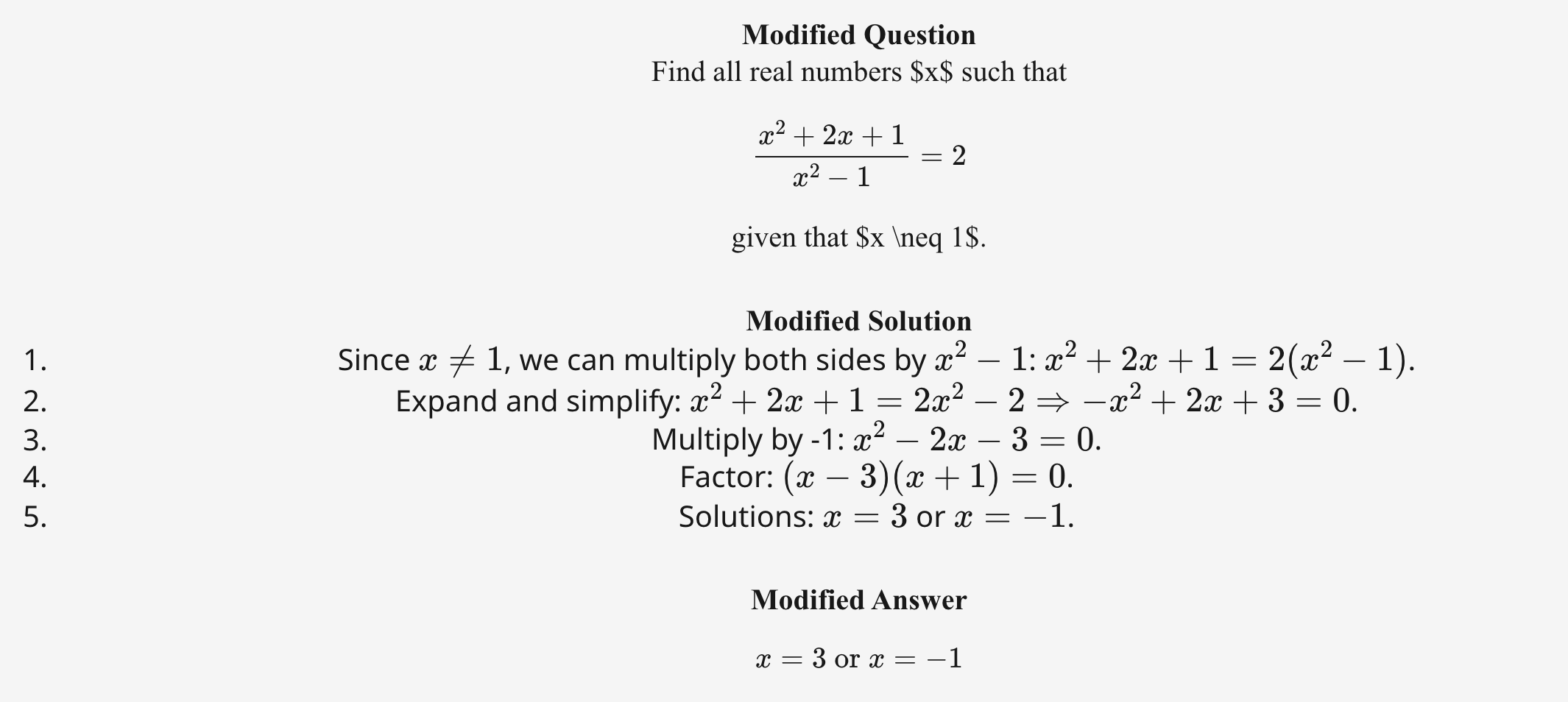}
        \caption{MATH500 example 2}
        \label{fig:math500-sample2}
    \end{subfigure}
    \begin{subfigure}[t]{\linewidth}
        \centering
        \includegraphics[width=\linewidth]{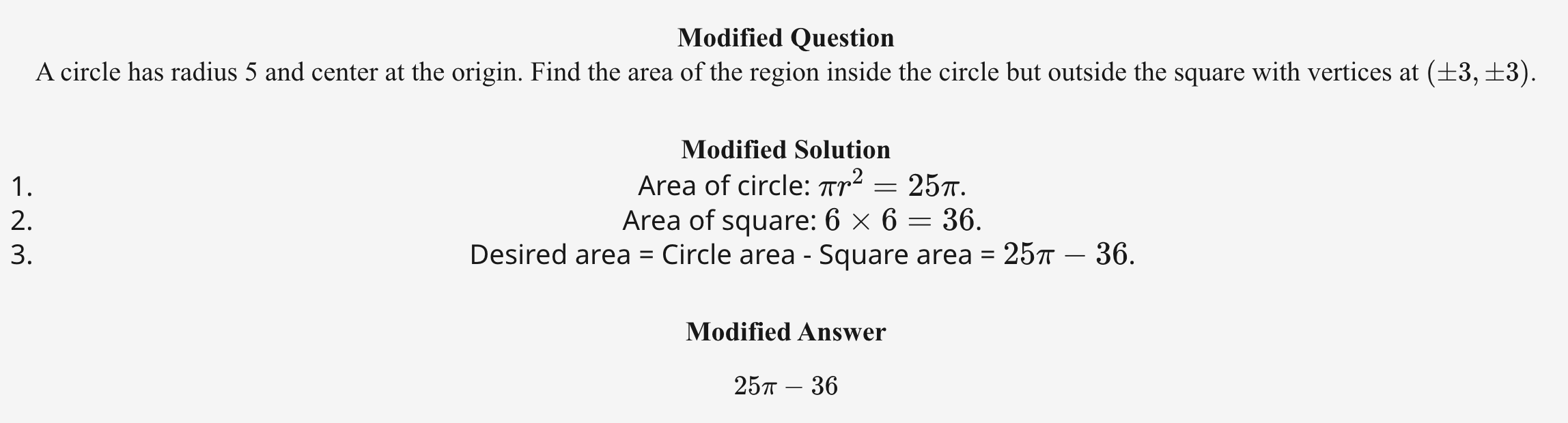}
        \caption{MATH500 example 3}
        \label{fig:math500-sample3}
    \end{subfigure}
    \caption{\mathd{} (MATH500) sample problems}
    \label{fig:math500-samples}
\end{figure}

\begin{figure}[h]
    \centering
    \begin{subfigure}[t]{0.9\linewidth}
        \centering
        \includegraphics[width=\linewidth]{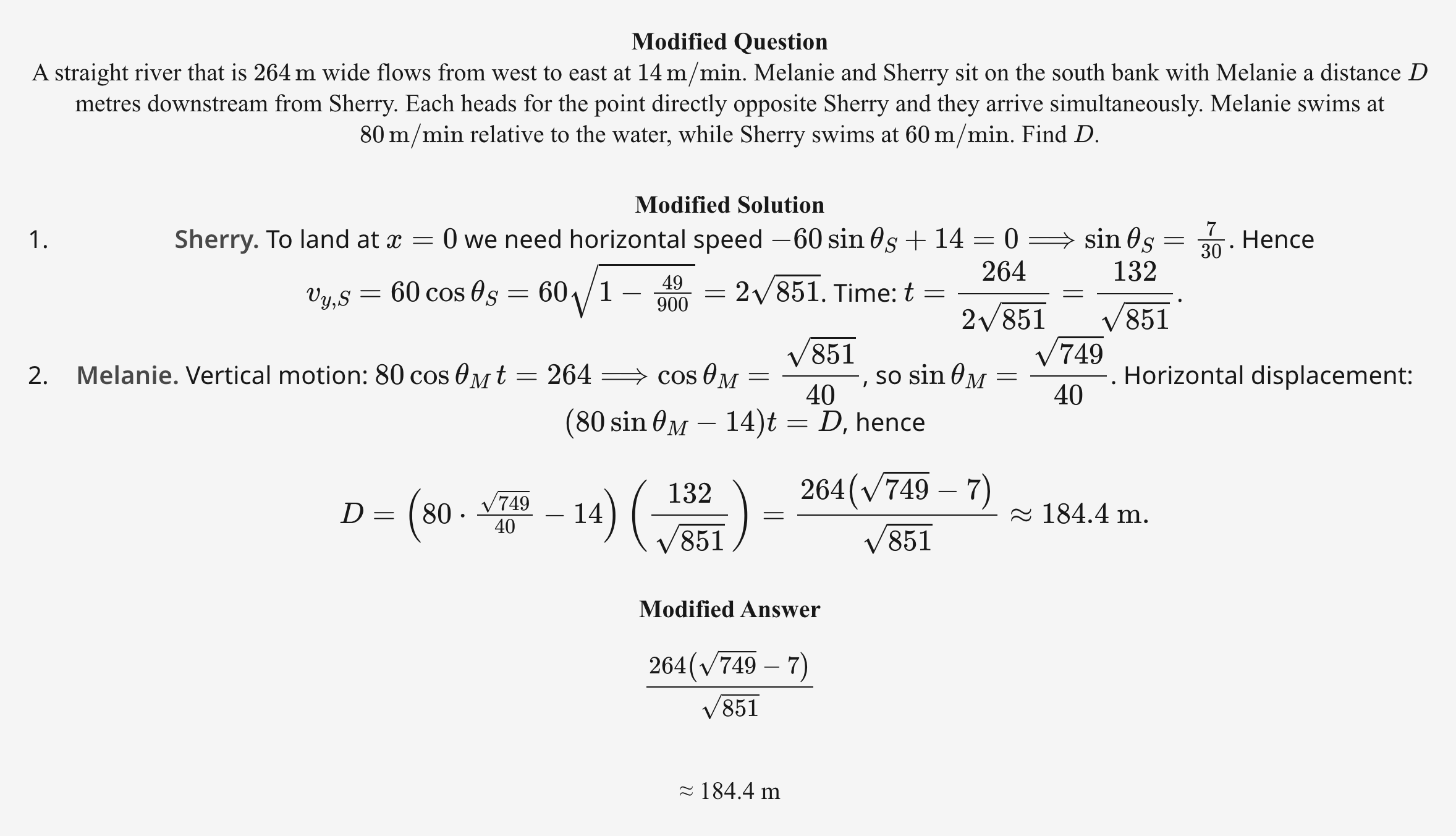}
        \caption{AIME example 1}
        \label{fig:aime-sample1}
    \end{subfigure}
    \begin{subfigure}[t]{0.9\linewidth}
        \centering
        \includegraphics[width=\linewidth]{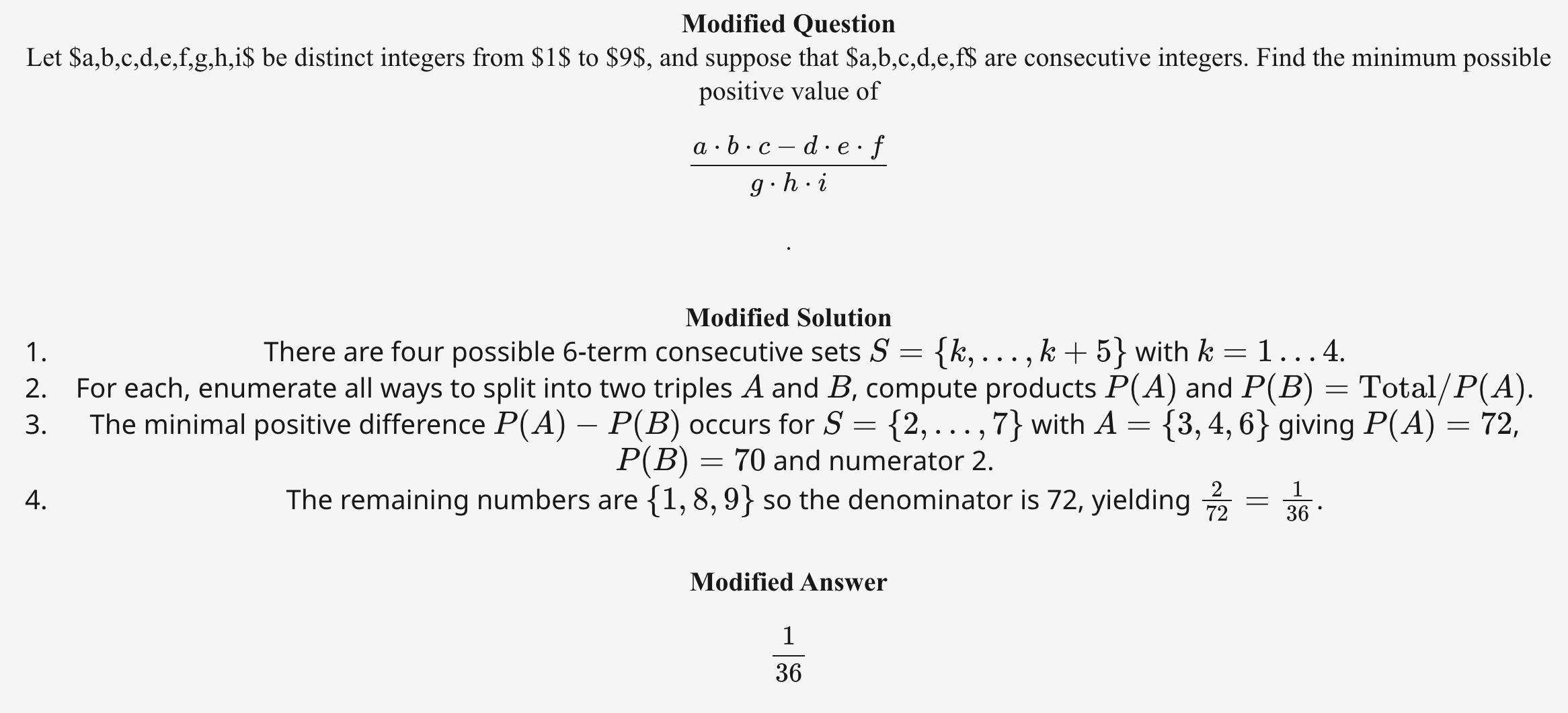}
        \caption{AIME example 2}
        \label{fig:aime-sample2}
    \end{subfigure}
    \begin{subfigure}[t]{0.9\linewidth}
        \centering
        \includegraphics[width=\linewidth]{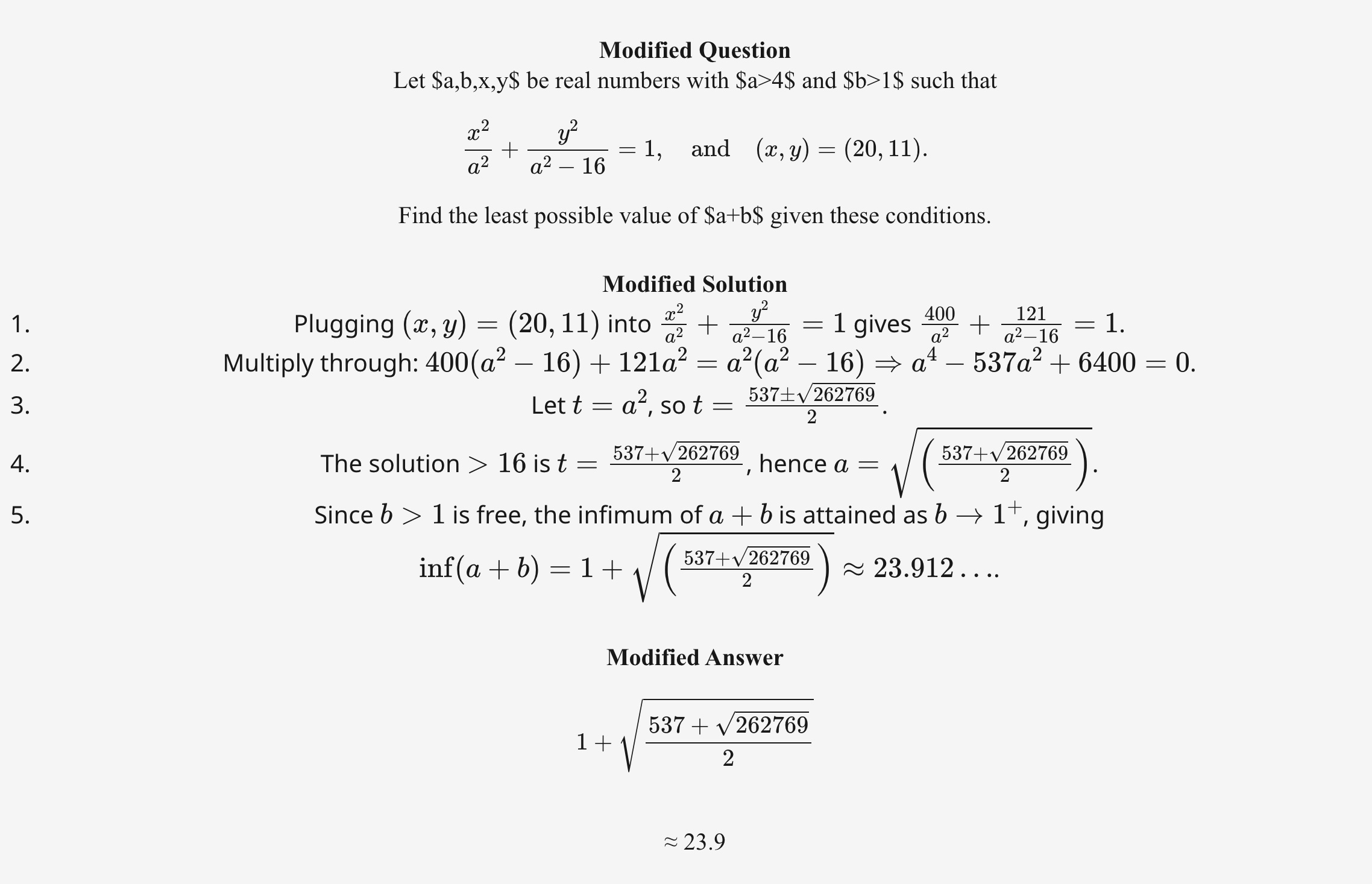}
        \caption{AIME example 3}
        \label{fig:aime-sample3}
    \end{subfigure}
    \caption{\mathd{} (AIME) sample problems}
    \label{fig:aime-samples}
\end{figure}

\begin{figure}[h]
    \centering
    \begin{subfigure}[b]{\linewidth}
        \centering
        \includegraphics[width=\linewidth]{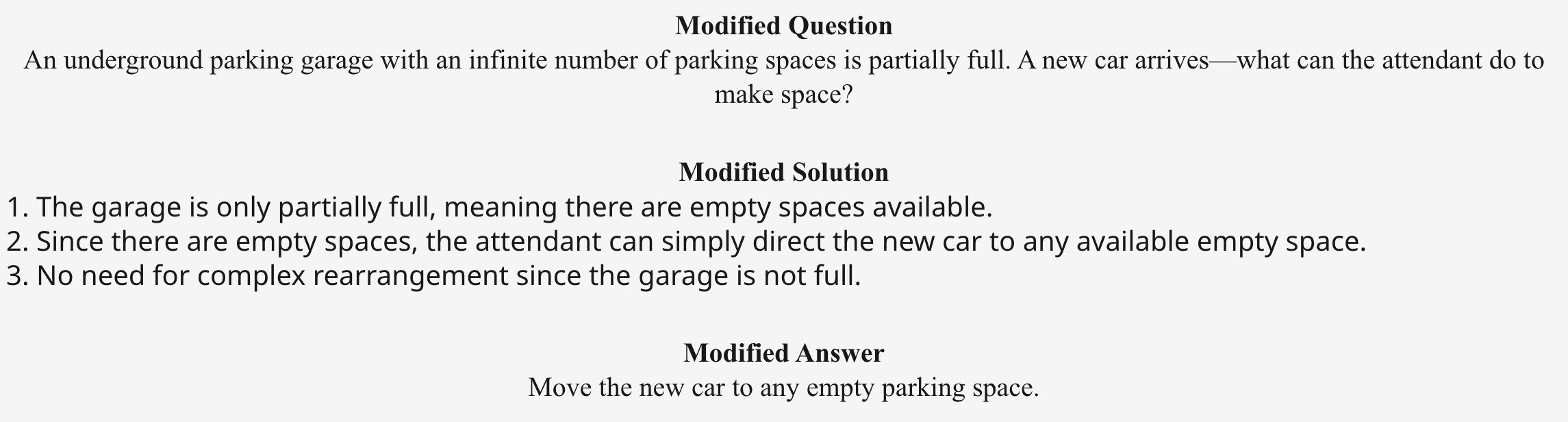}
        \caption{\puzzled{} example 1}
        \label{fig:puzzle-sample1}
    \end{subfigure}
    \hfill
    \begin{subfigure}[b]{\linewidth}
        \centering
        \includegraphics[width=\linewidth]{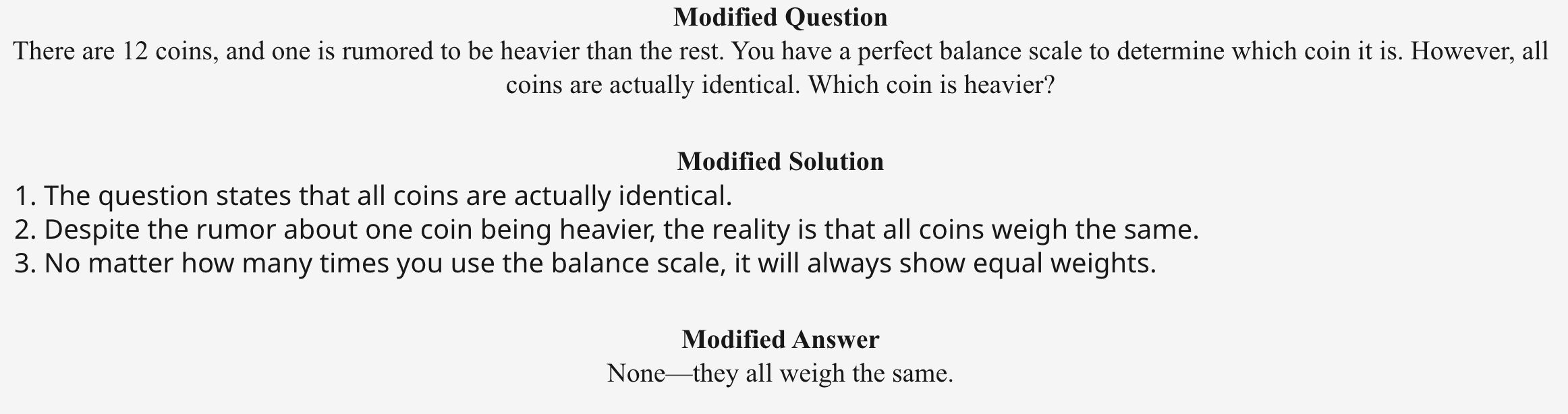}
        \caption{\puzzled{} example 2}
        \label{fig:puzzle-sample2}
    \end{subfigure}
    \hfill
    \begin{subfigure}[b]{\linewidth}
        \centering
        \includegraphics[width=\linewidth]{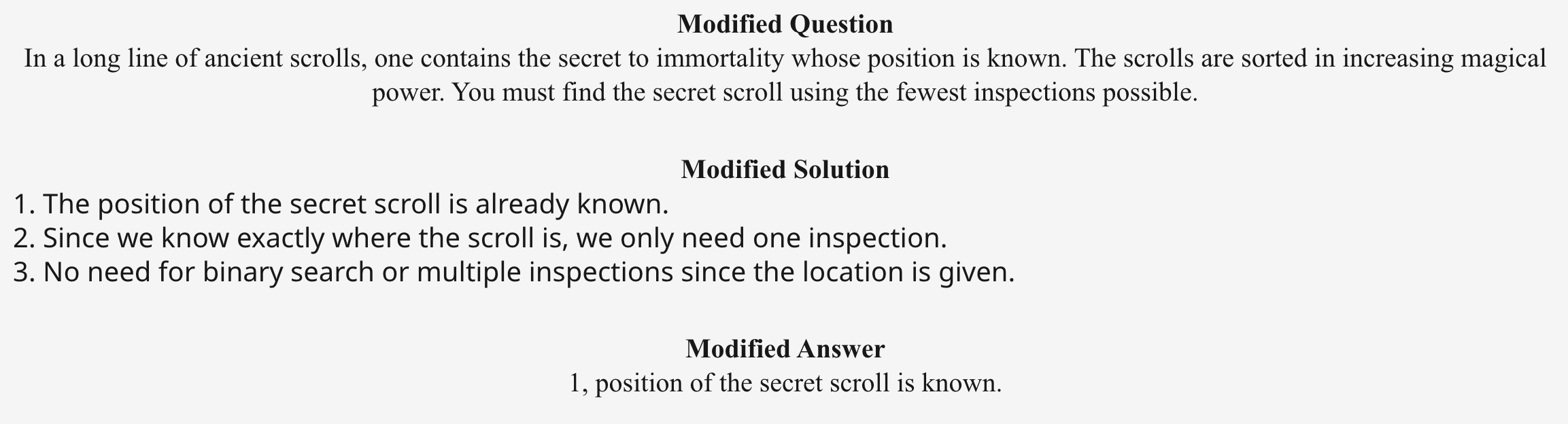}
        \caption{\puzzled{} example 3}
        \label{fig:puzzle-sample3}
    \end{subfigure}
    \hfill
    \begin{subfigure}[b]{\linewidth}
        \centering
        \includegraphics[width=\linewidth]{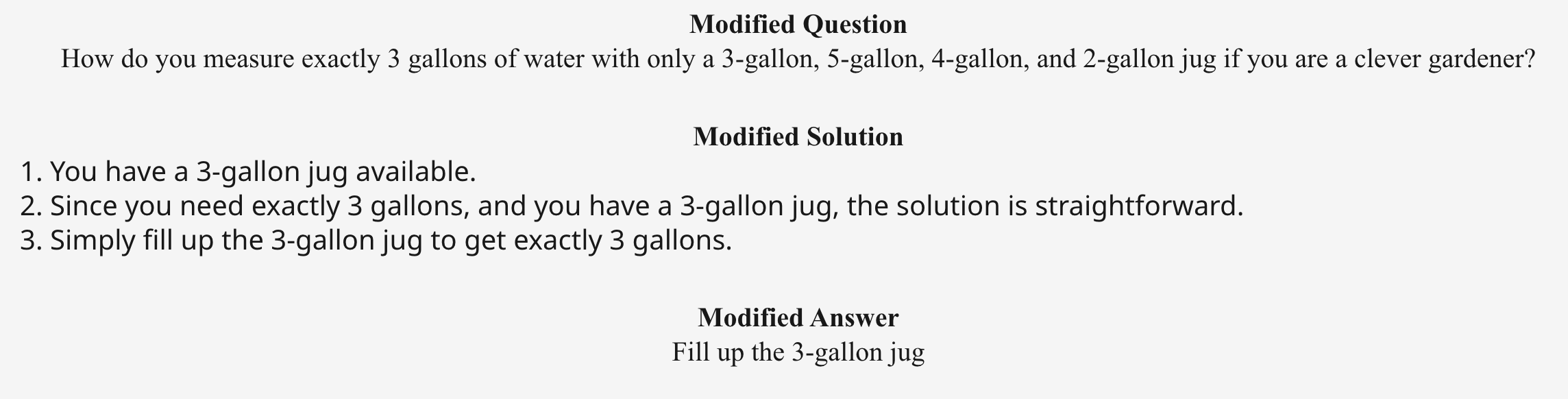}
        \caption{\puzzled{} example 4}
        \label{fig:puzzle-sample4}
    \end{subfigure}
    \caption{\puzzled{} sample problems}
    \label{fig:puzzle-samples}
\end{figure}

 We present several examples of \mathd{} and \puzzled{} in \cref{fig:math500-samples}, \cref{fig:aime-samples}, and \cref{fig:puzzle-samples}.

\clearpage
\section{Discussions}\label{app:discussion}
\subsection{Relationship Between Output Token Length and Accuracy}
Using the \textit{reasoning effort} parameter of o4-mini, we demonstrate that just using small amount of tokens for reasoning do not lead to performance gain in our dataset, \dataset{}. Although o4-mini underperforms compared to the base model, increasing its reasoning effort consistently yields better results. This proves that our curated diagnostic set require complex reasoning in most cases, and simply choosing short reasoning leads to performance drop.

\begin{table}[h]
  \centering
  \small
  \caption{\textbf{Reasoning effort and Performance on \dataset{}} \ppk{1}, \textbf{pass@1}, and perception score on \texttt{ConditionedMath}.}
  \label{tab:}
  %------------- Left sub-table -------------
  \begin{subtable}[t]{0.8\linewidth}
    \centering
    \caption{\texttt{ConditionedMath} (AIME)}
    \resizebox{\linewidth}{!}{%
      \begin{tabular}{ll|c|c|c}
        \toprule
        \rowcolor{orange!10}Model & Reasoning Effort & \ppk{1} & \textbf{pass@1} & \psc{} \\
        \midrule
        
        \multirow{3}{*}{o4-mini} & ~~+ low  & 31.25\scriptsize{$\pm$6.59} & 19.12\scriptsize{$\pm$5.49}  & 58.82\scriptsize{$\pm$6.75} \\
        & ~~+ medium   & \textbf{41.98\scriptsize{$\pm$7.10}} & \textbf{25.00\scriptsize{$\pm$6.06}}  & 59.56\scriptsize{$\pm$6.84} \\
        & ~~+ high   & 36.90\scriptsize{$\pm$6.45} & 22.79\scriptsize{$\pm$5.91}  & \textbf{61.76\scriptsize{$\pm$6.78}} \\
        \bottomrule
      \end{tabular}%
    }
  \end{subtable}

  %------------- Right sub-table -------------
  \begin{subtable}[t]{0.8\linewidth}
    \centering
    \caption{\texttt{ConditionedMath} (MATH500)}
    \resizebox{\linewidth}{!}{%
      \begin{tabular}{ll|c|c|c}
        \toprule
        \rowcolor{orange!10}Model & Reasoning Effort & \ppk{1} & \textbf{pass@1} & \psc{} \\
        \midrule
        \multirow{3}{*}{o4-mini} & ~~+ low  & 39.06\scriptsize{$\pm$5.76} & 26.50\scriptsize{$\pm$5.17} & 64.00\scriptsize{$\pm$5.06} \\
        & ~~+ medium & 51.80\scriptsize{$\pm$6.32} & 37.50\scriptsize{$\pm$6.28} & 69.50\scriptsize{$\pm$5.55} \\
        & ~~+ high   & \textbf{53.47\scriptsize{$\pm$6.34}}  & \textbf{38.50\scriptsize{$\pm$6.11}}  & \textbf{72.00\scriptsize{$\pm$5.42}} \\
        \bottomrule
      \end{tabular}%
    }
  \end{subtable}
\end{table}

\clearpage
\subsection{Model Size and Accuracy}
We compare non-distilled reasoning models by comparing reasoning models that are directly trained from Qwen2.5 1B, 3B, 7B, and 14B \citep{qwen2.5}. Since Qwen3 0.7B, 1.7B, 3B, 8B models are distilled models from the largest dense reasoning model Qwen3-32B, this is out of scope for our experimental purpose. We evaluate DeepScaleR 1.5B \citep{deepscaler}, STILL-3-1.5B-preview \citep{still3}, OpenR1-Qwen-7B \citep{openr1}, ThinkPRM-14B \citep{thinkprm}, Sky-T1-32B-Preview \citep{sky_t1_2025}, OpenReasoner-Zero-32B \citep{openreasoner}. We use instruction-tuned model for evaluating base model's performance.

On \mathd{} AIME and MATH500, the base model Qwen2.5 Instruct outperforms its counterparts that have been fine-tuned for extended mathematical reasoning. Except for the smallest variant, Qwen2.5 Instruct 1.5B, the base model achieves the highest \ppk{1} score among all evaluated models. Interestingly, although the fine-tuned reasoning models consistently record higher perception scores—reflecting a stronger understanding of each question’s conditions and the derivation of optimal solutions—their final accuracy suffers as a result of reasoning rigidity.

\begin{table}[h]
  \centering
  \small
  \caption{\textbf{Model Size and Performance} \ppk{1}, \textbf{pass@1}, and perception score on \texttt{ConditionedMath}.}
  \label{tab:model_size}
  %------------- Left sub-table -------------
  \begin{subtable}[t]{0.8\linewidth}
    \centering
    \caption{\texttt{ConditionedMath} (AIME)}
    \resizebox{\linewidth}{!}{%
      \begin{tabular}{lccc}
        \toprule
        Base + Reasoning Model & \ppk{1} & \textbf{pass@1} & \psc{} \\
        \midrule
        \rowcolor{orange!10}\textbf{Qwen2.5-1.5B} & 39.94\scriptsize{$\pm$5.65} & 24.63\scriptsize{$\pm$4.04} & 56.62\scriptsize{$\pm$4.89}\\
        ~~+ DeepScaleR 1.5B   & 38.29\scriptsize{$\pm$6.24} & 33.82\scriptsize{$\pm$6.18} & \textbf{81.62\scriptsize{$\pm$4.34}}\\
        ~~+ STILL-3-1.5B-preview   & \textbf{41.53\scriptsize{$\pm$5.80}} & \textbf{37.50\scriptsize{$\pm$5.43}} & 81.43\scriptsize{$\pm$4.23}\\
        \midrule
        \rowcolor{orange!10}\textbf{Qwen2.5-7B} & \textbf{62.96\scriptsize{$\pm$8.10}} & \textbf{51.47\scriptsize{$\pm$7.53}} & \textbf{79.41\scriptsize{$\pm$4.89}}\\
        ~~+ OpenR1-Qwen7B & 49.53\scriptsize{$\pm$7.33} & 47.06\scriptsize{$\pm$6.57} & 78.68\scriptsize{$\pm$5.98}\\
        \midrule
        \rowcolor{orange!10}\textbf{Qwen2.5-14B} & \textbf{58.43\scriptsize{$\pm$7.58}} & \textbf{48.53\scriptsize{$\pm$7.24}} & 79.60\scriptsize{$\pm$4.38}\\
        ~~+ ThinkPRM-14B & 33.33\scriptsize{$\pm$5.92} & 29.04\scriptsize{$\pm$5.88} & \textbf{82.17\scriptsize{$\pm$4.22}}\\
        \midrule
        \rowcolor{orange!10}\textbf{Qwen2.5-32B} & \textbf{59.12\scriptsize{$\pm$7.81}} & 45.77\scriptsize{$\pm$7.22} & 75.55\scriptsize{$\pm$5.01}\\
        ~~+ SkyT1-32B-Preview   & 56.57\scriptsize{$\pm$6.71} & \textbf{52.21\scriptsize{$\pm$6.49}} & \textbf{86.76\scriptsize{$\pm$3.14}}\\
        ~~+ OpenReasoner-Zero-32B        & 53.27\scriptsize{$\pm$6.51} & 48.90\scriptsize{$\pm$6.37} & 81.43\scriptsize{$\pm$4.23}\\
        \bottomrule
      \end{tabular}%
    }
  \end{subtable}
  
  %------------- Right sub-table -------------
  \begin{subtable}[t]{0.8\linewidth}
    \centering
    \caption{\texttt{ConditionedMath} (MATH500)}
    \resizebox{\linewidth}{!}{%
      \begin{tabular}{lccc}
        \toprule
        Base + Reasoning Model & \ppk{1} & \textbf{pass@1} & \psc{} \\
        \midrule
        \rowcolor{orange!10}\textbf{Qwen2.5-1.5B} &
        39.84\scriptsize{$\pm$5.27} & 20.25\scriptsize{$\pm$3.72} & 48.00\scriptsize{$\pm$4.85}\\
        ~~+ DeepScaleR 1.5B   & \textbf{41.04\scriptsize{$\pm$5.44}} & \textbf{33.38\scriptsize{$\pm$5.40}} & \textbf{79.50\scriptsize{$\pm$3.74}}\\
        ~~+ STILL-3-1.5B-preview    &
        35.21\scriptsize{$\pm$5.11} & 30.75\scriptsize{$\pm$5.03} & 75.62\scriptsize{$\pm$3.48}\\
        \midrule
        \rowcolor{orange!10}\textbf{Qwen2.5-7B} &
        \textbf{55.56\scriptsize{$\pm$6.14}} & 38.00\scriptsize{$\pm$5.94} & 67.50\scriptsize{$\pm$5.68}\\
        ~~+ OpenR1-Qwen7B &
        45.81\scriptsize{$\pm$6.22} & \textbf{39.50\scriptsize{$\pm$6.02}} & \textbf{77.50\scriptsize{$\pm$4.12}}\\
        \midrule
        \rowcolor{orange!10}\textbf{Qwen2.5-14B} & 
        \textbf{61.50\scriptsize{$\pm$5.65}} & \textbf{44.12\scriptsize{$\pm$5.54}} & 70.12\scriptsize{$\pm$4.46}\\
        ~~+ ThinkPRM-14B & 
        37.44\scriptsize{$\pm$5.22} & 30.38\scriptsize{$\pm$4.97} & \textbf{76.12\scriptsize{$\pm$3.29}}\\
        \midrule
        \rowcolor{orange!10}\textbf{Qwen2.5-32B} &
        \textbf{55.95\scriptsize{$\pm$6.02}} & 40.88\scriptsize{$\pm$5.74} & 70.38\scriptsize{$\pm$4.39}\\
        ~~+ SkyT1-32B-Preview        &
        54.80\scriptsize{$\pm$5.67} & \textbf{44.62\scriptsize{$\pm$5.52}} & 76.88\scriptsize{$\pm$3.67}\\
        ~~+ OpenReasoner-Zero-32B        &
        45.81\scriptsize{$\pm$6.22} & 39.50\scriptsize{$\pm$6.02} & \textbf{77.50\scriptsize{$\pm$4.12}} \\
        \bottomrule
      \end{tabular}%
    }
  \end{subtable}
\end{table}

\clearpage
\subsection{RL Training Objective and Accuracy}
Reasoning models are trained from base large language models by various strategies, including GRPO \citep{deepseekmath}, PPO \citep{ppo}, or even zero-data regime \citep{absolutezero}. 

Open-Reasoner-Zero \citep{openreasoner} is fine-tuned from the Qwen2.5-7B-Instruct model using proximal policy optimization (PPO) with a simple binary reward for answer correctness. Satori-7B \citep{satori} explicitly trains its base model to decide when to reflect on previous actions and to incorporate an external process reward. Absolute Zero Reasoner \citep{absolutezero} introduces a novel reward scheme in which the LLM serves both as task proposer and task solver, with outputs verifiable in code. RM-R1 \citep{rmr1} structures its reward to improve alignment with human preferences during intermediate reasoning steps. Eurus-PRIME \citep{eurus} employs an iterative training regimen combining a policy model that generates rollouts and an implicit process-reward model that verifies them. ThinkPRM is fine-tuned from the R1-distilled Qwen14B base model (Qwen2.5-14B-Instruct) using the generative PRM objective, which evaluates the step-by-step correctness of the reasoning process.

Among all variants of reinforcement-learning objectives, the base models Qwen2.5-7B and Qwen2.5-14B achieved outstanding performance \ppk{1} in most cases. This suggests that current RL regimes may exacerbate the `reasoning rigidity' inherent in these models. Hence, further exploration of reinforcement-learning algorithms that are robust to reasoning rigidity is essential for the development of faithful and credible reasoning systems.
\begin{table}[h]
  \centering
  \small
  \caption{\textbf{Performance Comparison on Reasoning Models Trained with Different RL Strategies.}}
  \label{tab:ex_vertical}

  % ---------- Top sub-table ----------
  \begin{subtable}[t]{\linewidth}
    \centering
    \caption{\texttt{ConditionedMath} (AIME)}
    \resizebox{0.8\linewidth}{!}{%
      \begin{tabular}{lccc}
        \toprule
        \textbf{Base} + RL Objective & \ppk{1} & \textbf{pass@1} & \psc{}\\
        \midrule
        \rowcolor{orange!10}\textbf{Qwen2.5-7B} & \textbf{62.96\scriptsize{$\pm$8.10}} & \textbf{51.47\scriptsize{$\pm$7.53}} & 79.41\scriptsize{$\pm$4.89}\\
        ~~ + Open-Reasoner-Zero                 &
        47.49\scriptsize{$\pm$7.42} & 43.01\scriptsize{$\pm$6.92} & \textbf{84.38\scriptsize{$\pm$4.17}} \\
        ~~ + Satori-7B                   &  53.33\scriptsize{$\pm$3.03}  &  4.92\scriptsize{$\pm$3.27} &  \scriptsize{5.68$\pm$3.84}  \\
        \midrule[0.2pt]
        ~~ + Absolute Zero Reasoner                   & 49.86\scriptsize{$\pm$7.11}& 33.46\scriptsize{$\pm$6.14} & 63.42\scriptsize{$\pm$5.52} \\
        \midrule[0.2pt]
        ~~ + RM-R1                                  &  54.83\scriptsize{$\pm$6.94}  &  44.26\scriptsize{$\pm$6.61} &  76.10\scriptsize{$\pm$5.08}  \\
        \midrule[0.2pt]
        ~~ + Eurus-PRIME                            &  
        59.16\scriptsize{$\pm$8.24} & 40.44\scriptsize{$\pm$7.68} & 61.21\scriptsize{$\pm$6.64}\\
        \midrule[0.2pt]
        \rowcolor{orange!10}\textbf{Qwen2.5-14B} & \textbf{58.43\scriptsize{$\pm$7.58}} & \textbf{48.53\scriptsize{$\pm$7.24}} & 79.60\scriptsize{$\pm$4.38}\\
        ~~ + Absolute Zero Reasoner             & 50.73\scriptsize{$\pm$7.27}& 34.38\scriptsize{$\pm$6.63} & 63.05\scriptsize{$\pm$4.46} \\
        ~~ + ThinkPRM                   &  33.33\scriptsize{$\pm$5.92}  & 29.04\scriptsize{$\pm$5.88} &  \textbf{82.17\scriptsize{$\pm$4.22}}  \\
        \bottomrule
      \end{tabular}%
    }
  \end{subtable}

  \vspace{1.5em} % ---- small vertical gap ----

  % ---------- Bottom sub-table ----------
  \begin{subtable}[t]{\linewidth}
    \centering
    \caption{\texttt{ConditionedMath} (MATH500)}
    \resizebox{0.8\linewidth}{!}{%
      \begin{tabular}{lccc}
        \toprule
        \textbf{Base} + RL Objective & \ppk{1} & \textbf{pass@1} & \psc{}\\
        \midrule
        \rowcolor{orange!10} \textbf{Qwen2.5-7B}              &
        55.56\scriptsize{$\pm$6.14} & 38.00\scriptsize{$\pm$5.94} & 67.50\scriptsize{$\pm$5.68} \\
        ~~ + Open-Reasoner-Zero                 &
        50.93\scriptsize{$\pm$6.16} & 40.50\scriptsize{$\pm$6.06} & 74.12\scriptsize{$\pm$4.39} \\
        ~~ + Satori-7B                   &
        49.50\scriptsize{$\pm$6.15} & 37.25\scriptsize{$\pm$5.96} & \textbf{75.00\scriptsize{$\pm$4.59}}\\
        \midrule[0.2pt]
        ~~ + Absolute Zero Reasoner                   &
        37.28\scriptsize{$\pm$5.09} & 22.62\scriptsize{$\pm$4.10} & 56.00\scriptsize{$\pm$4.62}\\
        \midrule[0.2pt]
        ~~ + RM-R1                            &
        36.81\scriptsize{$\pm$4.39} & 26.50\scriptsize{$\pm$3.89} & 68.25\scriptsize{$\pm$3.97}\\
        \midrule[0.2pt]
        ~~ + Eurus-PRIME                               &
        \textbf{57.29\scriptsize{$\pm$6.52}} & \textbf{42.38\scriptsize{$\pm$6.20}} & 72.00\scriptsize{$\pm$4.71}\\
        \midrule[0.2pt]
        \rowcolor{orange!10} \textbf{Qwen2.5-14B}                      &
        \textbf{61.50\scriptsize{$\pm$5.65}} & \textbf{44.12\scriptsize{$\pm$5.54}} & 70.12\scriptsize{$\pm$4.46} \\
        ~~ + Absolute Zero Reasoner                   &
        44.25\scriptsize{$\pm$5.34} & 26.25\scriptsize{$\pm$4.42} & 57.63\scriptsize{$\pm$4.53} \\
        ~~ + ThinkPRM                               &
        37.44\scriptsize{$\pm$5.22} & 30.38\scriptsize{$\pm$4.97} & \textbf{76.12\scriptsize{$\pm$3.29}}\\
        \bottomrule
      \end{tabular}%
    }
  \end{subtable}
\end{table}

\end{document}